\PassOptionsToPackage{svgnames}{xcolor}
\documentclass{article} 
\input{setting.tex}

\title{\textbf{Causal Regularization}%
\date{}
\author{
Mohammad Taha Bahadori$^1$,~ Krzysztof Chalupka$^2$,~ Edward Choi$^1$,\\
Robert Chen$^1$, ~Walter F. Stewart$^3$,~ \& Jimeng Sun$^1$\\
$^1$Georgia Institute of Technology, \; $^2$California Institute of Technology, \; $^3$Sutter Health \\
} %
}

\begin{document}
\maketitle
\begin{abstract}
In application domains such as healthcare, we want accurate predictive models that are also causally interpretable. In pursuit of such models, we propose a causal regularizer to steer predictive models towards causally-interpretable solutions and theoretically study its properties. In a large-scale analysis of Electronic Health Records (EHR), our causally-regularized model outperforms its $L_1$-regularized counterpart in causal accuracy and is competitive in predictive performance. We perform non-linear causality analysis by causally regularizing a special neural network architecture.
We also show that the proposed causal regularizer can be used together with neural representation learning algorithms to yield up to $20\%$ improvement over multilayer perceptron in detecting multivariate causation, a situation common in healthcare, where many causal factors should occur simultaneously to have an effect on the target variable.

\end{abstract}

\section{Introduction}
\label{sec:intro}
In domains such as healthcare, genomics or social science there is high demand for data analysis that reveals \emph{causal} relationships between independent and target variables. For example, doctors not only want models that accurately predict the status of patients, but also want to identify the factors that can improve it.
The distinction between prediction and causation has at times been subject to debate in statistics and machine learning \citep{breiman2001statistical,shmueli2010explain}. While machine learning has focused mostly on prediction tasks, in many scientific domains pure prediction without considering the underlying causal mechanisms is considered unscientific \citep{shmueli2010explain}.  In this work, we propose a {\it causal regularizer} that balances causal interpretability and predictive power. 

\begin{figure}[t]
\centering
\includegraphics[scale=0.25]{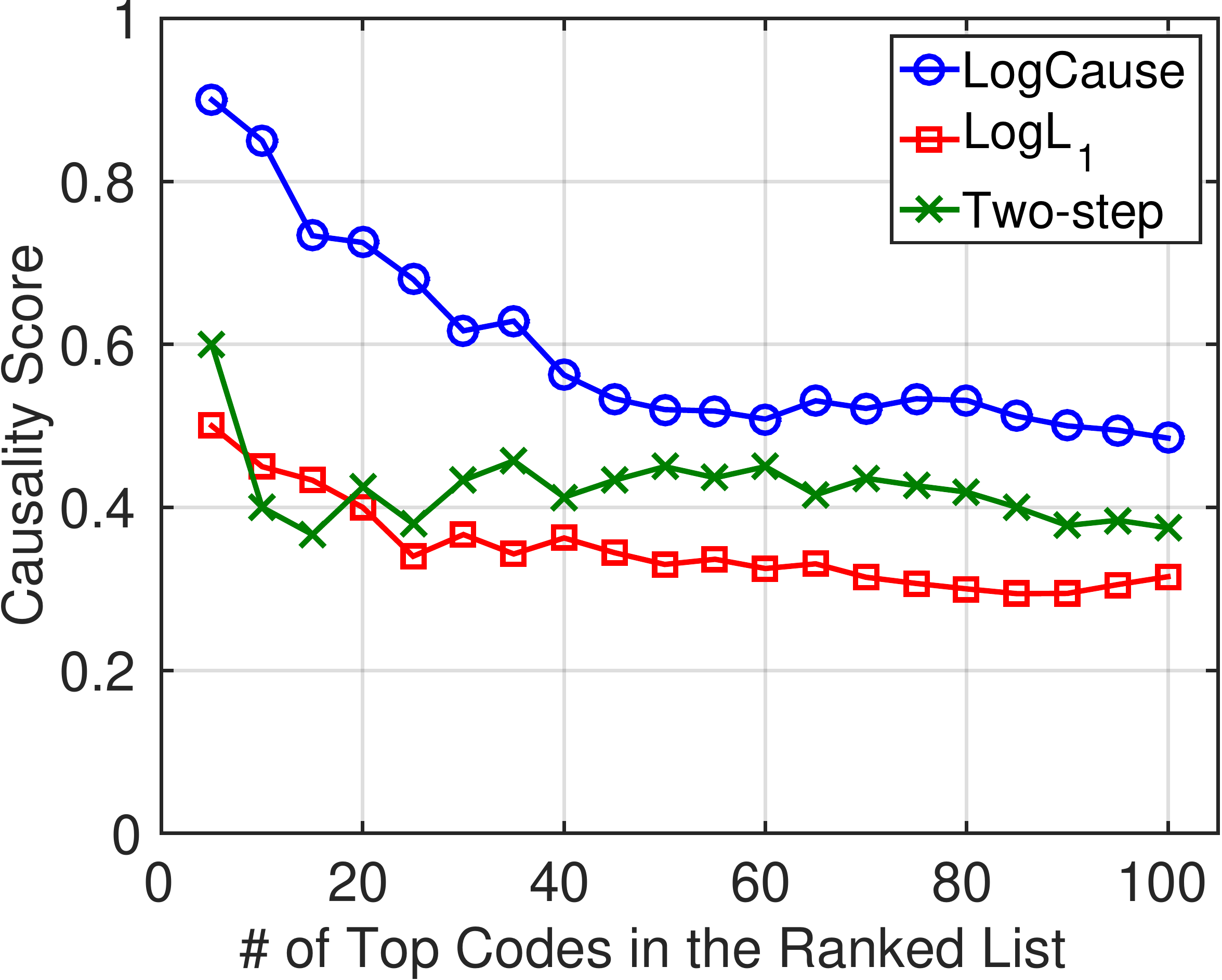}
        \caption{The proposed causal regularizer achieves significantly higher causality score, computed using ground truth causality. We compute the score for top $k$ codes in the ranked list reported by three algorithms. The causal regularizer is also competitive in predictive performance, see Section \ref{sec:exp} for more details.}
        \label{fig:causality_intro}
\end{figure}


We use the counterfactual causality framework \citep{pearl2009causality}, in which one random variable $X$ (e.g. red wine consumption) \emph{causes} another variable $Y$ (i.e., reduction in risk of heart attack) denoted as $X \rightarrow Y$ if \textit{experimental testing} of $X$ would be proven to change the distribution of $Y$ \citep{spirtes2010introduction}. But there may also be competing explanations of the observed correlation between $X$ and $Y$ because of confounding (e.g., people of high socio-economic status tend to drink more wine and this is related to other lifestyle factors that cause a reduction in heart attack) that need to be reconciled in assessing the likelihood that $X \to Y$ is true. 
Causal analytic methods can be used to prioritize what warrants testing in clinical trials among a diversity of hypotheses or as primary evidence if  controlled trials are not feasible or desireable (e.g., climate science or health).  In healthcare, in particular,  it is common that an ensemble  of many causal factors needs to occur simultaneously to have an effect on the target variable, a phenomenon we will call \textit{multivariate causation}. Scalable methods are needed to explore the exponential combinations of the independent variables and different transformations in order to detect multivariate causal relationships.

Methods for discovering causal relationships among multiple variables from observational data \citep{chickering2002optimal,kalisch2007estimating,colombo2012learning} 
are largely based on the principle that any given set of causal relationships among multiple variables leaves well-defined marks in the joint distribution of the variables. However, when these methods are used for causal variable selection \citep{guyon2007causal,cawley2008causal,bontempi2010causal,sun2015using}, the process becomes very sensitive to small changes in the joint distribution of variables and may exclude many causal variables due to noise or selection bias in the data.

Our main idea is to design a causal regularizer to control the complexity of the statistical models and at the same time favor causal explanations. Compared to the two step procedure of causal variable selection followed by a multivariate regression/classification, the proposed approach performs joint causal variable selection and prediction, thus avoiding the statistically sensitive thresholding of the causality scores in the causal variable selection step. It allows few dependencies that cannot be explained via causation to still be included in the model, relaxing the variable selection procedure. Our technical contributions are as follows: 

\begin{enumerate}[leftmargin=5mm]
\item We use causality detectors to construct a {\bf causal regularizer} that can guide predictive models towards learning causal relationships between the independent and target variables. We theoretically quantify the impact of the accuracy of the causality detector on the causal accuracy of the regularized models.
\item We propose a new non-linear predictive model regularized by our causal regularizer, which allows \textbf{causally interpretable neural networks}.
\item Finally, we demonstrate that the proposed causal regularizer can be combined with neural representation learning techniques to efficiently detect \textbf{multivariate causal hypotheses}.
\end{enumerate}
The proposed framework scales linearly with the number of variables, as opposed to many previous causal methods. 

We applied the proposed algorithms to clinical predictive modeling problems using large EHR datasets: one on heart failure onset prediction and another on mortality prediction using the publicly available MIMIC III \citep{johnson2016mimic} dataset. Altogether, we analyzed the collective influence of 17,081 independent variables on heart failure and validated the results by having a clinical expert to manually review the findings in a blind setup. 
As shown in Figure \ref{fig:causality_intro}, our proposed causally-regularized algorithm significantly outperforms the baseline algorithms in causality detection performance. We show a similar boost in the causality score of the detected multivariate causal hypotheses. Finally, we show that the proposed algorithms are also competitive in predictive performance on both datasets.

\section{Preliminaries on Causality Detection}
\label{sec:detector}

We begin with description of pairwise causal analysis ($X\to Y$) of a single independent variable $X$ on the target variable $Y$ based on the independence of mechanisms (ICM) assumption and then extend the pairwise causality detector to perform multivariate causality analysis in the next section. While our proposed causal regularizer can be constructed using any causality detection algorithm, a review of the ICM based methods, as the state of art causality detection algorithms, is helpful because they are the baseline algorithms in the experiments.

\begin{figure}
    \centering
    \begin{subfigure}[b]{0.15\textwidth}
        \includegraphics[scale=0.5]{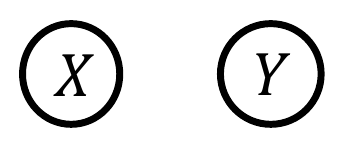}
        \caption{Independent}
        \label{fig:indep}
    \end{subfigure}
    ~ 
    \begin{subfigure}[b]{0.15\textwidth}
        \includegraphics[scale=0.5]{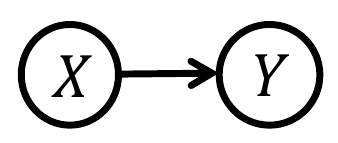}
        \caption{Direct}
        \label{fig:direct}
    \end{subfigure}
    ~
    \begin{subfigure}[b]{0.15\textwidth}
        \includegraphics[scale=0.5]{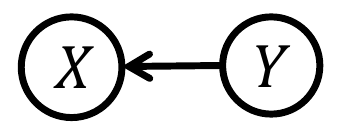}
        \caption{Reverse}
        \label{fig:reverse}
    \end{subfigure}
    ~
    \begin{subfigure}[b]{0.17\textwidth}
        \includegraphics[scale=0.5]{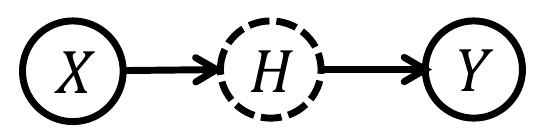}
        \caption{Indirect}
        \label{fig:indirect}
    \end{subfigure}
    ~ ~
    \begin{subfigure}[b]{0.2\textwidth}
        \includegraphics[scale=0.5]{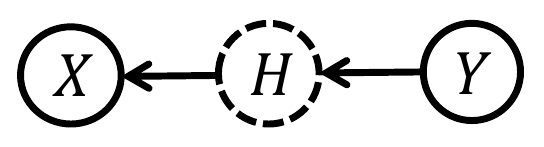}
        \caption{Indirect Reverse}
        \label{fig:inreverse}
    \end{subfigure}
    ~
    \begin{subfigure}[b]{0.17\textwidth}
        \includegraphics[width=\textwidth]{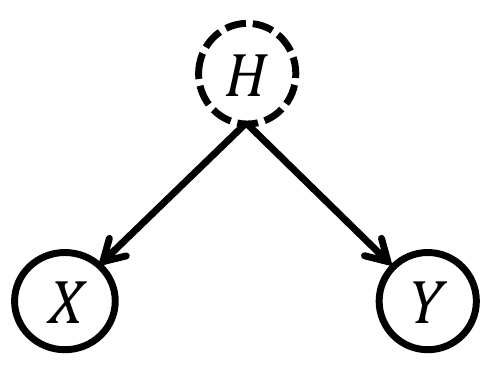}
        \caption{Confounded Correlation}
        \label{fig:conf}
    \end{subfigure}
    ~
    \begin{subfigure}[b]{0.17\textwidth}
        \includegraphics[width=\textwidth]{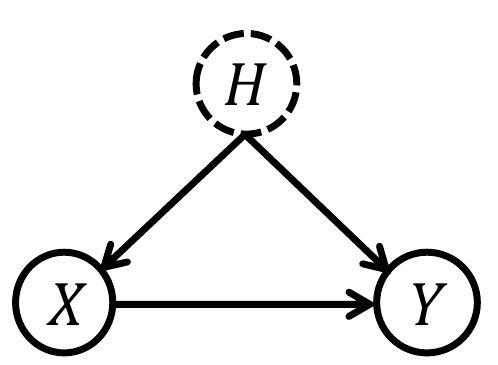}
        \caption{Confounded and Direct}
        \label{fig:conf_dir}
    \end{subfigure}
    ~
    \begin{subfigure}[b]{0.17\textwidth}
        \includegraphics[width=\textwidth]{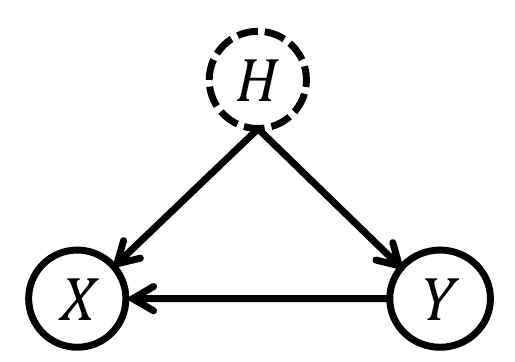}
        \caption{Confounded and Reverse}
        \label{fig:conf_rev}
    \end{subfigure}
    ~
    \begin{subfigure}[b]{0.17\textwidth}
        \includegraphics[width=\textwidth]{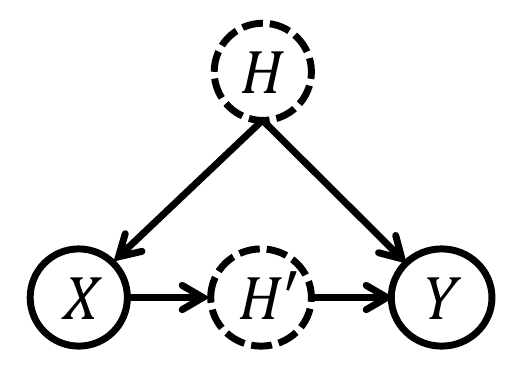}
        \caption{Confounded and Indirect}
        \label{fig:conf_indir}
    \end{subfigure}
    ~
    \begin{subfigure}[b]{0.17\textwidth}
        \includegraphics[width=\textwidth]{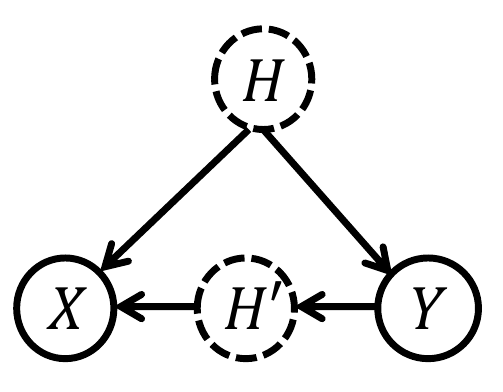}
        \caption{Confounded \& Indirect Reverse}
        \label{fig:conf_inrev}
    \end{subfigure}
    \caption{Common causal and anti-causal structures between two observed and one or more hidden variables. Under the algorithmic independence assumption, we can sample from the joint distribution of $X$ and $Y$ in each case and train a classifier to distinguish between these cases based on the (automatically learned) features of the joint distribution.}
    \label{fig:cases}
\end{figure}


We are interested in finding causal models  where $X$ causes $Y$, or $Y$  causes $X$, or the two are confounded based on joint distribution of $P(X,Y)$. However, the pairwise causality analysis is infeasible for arbitrary joint distributions. Thus, we need to resort to additional assumptions on the nature of the causal relationships. Recently several algorithms have been proposed that distinguish between the cause and effect based on the natural assumption that steps in the process that generates the data are independent from each other, see \citep{lemeire2006causal,janzing2012causal,daniusis2012inferring,lopezpaz2016,Chalupka2016,kocaoglu2016entropic} and the references therein. 
In this work, we follow \citep{lopez2016discovering,Chalupka2016} to describe this causality detection approach. In the next subsections, we describe our novel causal regularizer designed based on this causality detection approach and its application in non-linear causality analysis and multivariate causal hypothesis generation. 

\textbf{Conceptual description of the independence between the cause and the mechanism.}
ICM states that the two processes of generation of the cause and mapping from cause to effect are independent. In our case, we assume that when $X\to Y$ ($X$ causes $Y$), the probabilities  $\mathbb{P}(Y\mid X)$ and $\mathbb{P}(X)$ are generated by independent higher-level distribution functions. Thus, we do not put assumptions on the functional form of the causal relationships between the variables of interest. 
ICM conforms to the scientific idea of Uniformitarianism \citep{gould1965uniformitarianism} which, putting roughly, states that the laws of nature apply to all objects similarly. ICM can be described in both deterministic \citep{janzing2010causal} and probabilistic \citep{daniusis2012inferring} sense; this work mainly uses the probabilistic interpretation.

ICM can be used to generate samples from distributions that agree with the possible graphical models including two observed variables $X$ and $Y$ and an unobserved variable $H$ shown in Figure \ref{fig:cases}, by requiring that the probability functions in the factorization of the joint distribution are independent from each other. The hidden variables can represent the other observed variables, critical in design of the regularizer in the next subsection. \citet{Chalupka2016} developed an analytical likelihood ratio test that decides between the causal and anticausal cases (Figures \ref{fig:direct} and \ref{fig:reverse}). However, taking into account the confounded cases is analytically difficult. Nevertheless, it is possible to generate samples from the scenarios in Figure \ref{fig:cases} under the ICM and train a classifier to learn to choose the max likelihood causal structure given samples from the joint $P(\cs, \ef)$. This is the key idea of the causality detectors in \citep{lopez2016discovering,Chalupka2016} described in the rest of this section.

\textbf{Mathematical description of the causality detection algorithm.}
Formally, suppose we have $m$ variables $X_i$, each with dimensionality $d_i$. For each variable we observe a sample of size $n_i$ denoted by $S_i = \{(\mathbf{x}_{i,j}, y_j)\}_{j=1}^{n_i}$, where $y_j$ are observations of a common target variable $Y$. Let $\mathcal{S}$ denote the set of all such samples. For each sample $S_i$, we are interested in determining the binary label $\ell_i \in \{0,1\}$ which determines whether $X_i$ causes $Y$ or not. In fact, we are interested in the function approximation problem of learning the mapping $f: \mathcal{S}\mapsto\{0,1\}$.

Several approaches can learn such a mapping function. When $X$ and $Y$ are both discrete and finite, \citet{Chalupka2016} offer a means to construct the empirical joint distribution $\widehat{p}_i = \widehat{p}(X_i, Y)$ and train a supervised neural network mapping function $f(\widehat{p}_i)\to\ell_i$. \citet{lopez2016discovering} learn the representation $S_i \to \frac{1}{n_i}\sum_{j=1}^{n_i}\phi(\mathbf{x}_{i,j}, y_j)$ and a neural network $f'\left(\frac{1}{n_i}\sum_{j=1}^{n_i}\phi(\mathbf{x}_{i,j}, y_j)\right) \to \ell_i$, followed by training both the representation leaning function $\phi(\cdot, \cdot)$ and the classification network $f'(\cdot)$ in a joint and supervised way.

However, it is rare to have the true causal labels $\ell$ for training a causal detector. Rather, we generate synthetic datasets to represent the scenarios in Figure \ref{fig:cases} based on the ICM assumption. The overall procedure is to generate samples from distributions $p_{X,Y}$ that are one of the ten possible scenarios in Figure \ref{fig:cases}. We need to select distributions that impose a minimum number of restriction on the data and the synthetically-generated distributions have statistics as similar as possible to those of our true data of interest. For example, in our datasets, the independent variables $X$ are counts of the number of disease codes in patients' records (cf. Section \ref{sec:exp}). Thus, we sample $X$ from a mixture of appropriate distributions for count data:  the Zipf, Poisson, Uniform, and Bernoulli distributions. The hidden variable $H$ and the response variable $Y$ are sampled from the Dirichlet and Bernoulli distributions, respectively. Details of our sampling and training procedures are provided in Appendix \ref{sec:details_detector} and Algorithm \ref{blk:training} there.

\section{Methodology}
\label{sec:method}
Given the causality detector in Section \ref{sec:detector}, we propose the causal regularizer for linear models in \ref{sec:argue}. We demonstrate  in Section \ref{sec:hyp} using a non-linear deep neural networks regularized by our causal regularizer, we can learn non-linear causal relationships between the independent and target variables. Finally, we show that the causal regularizer can efficiently explore the space of multivariate causal hypotheses and extract meaningful candidates for causality analysis.


\subsection{The Causal Regularizer}
\label{sec:argue}


Using the causality detection methods in the previous section for causal variable selection \citep{guyon2007causal,cawley2008causal,bontempi2010causal,sun2015using} makes the variable selection process becomes very sensitive to small changes in the joint distribution of variables and may exclude many causal variables due to noise or selection bias in the data.
Ideally, if the ICM holds and if we had access to the true joint distributions and could discriminate between causal and non-causal variables with perfect accuracy, the two-step procedure would be sufficient. 
But observational  datasets are not usually an accurate representation of the true probabilistic generative process because of measurement error and selection bias, which can perturb the causality scores generated by the neural network causality detector.

For example, consider the two-step analysis process of first finding the variables that cause $Y$ from a list of variables $X_i$ for $i=1, \ldots, n$ and then performing a sparse multivariate regression on the selected variables to prioritize the selected variables. 
This procedure is sensitive because our causality detection algorithm might give soft scores such as $0.5+\varepsilon$ or $0.5-\varepsilon$ to two variables $X_1$ and $X_2$, respectively. These soft-scores can be interpreted as the probability that each variable is a cause of $Y$. If we use the two-step procedure, we will include $X_1$ in the regression model but not $X_2$. However, $X_2$ could possibly contribute more to the predictive performance in presence of other variables in the multivariate regression. In other words, any hard cut-off for the purpose of two-step causal variable selection and regression will pose the question of ``what should be the best cut-off threshold?''

Instead, we propose a causally regularized regression approach, where this trade-off is performed smoothly via a regularization parameter. We select variables that are both causal with high probability and also significantly predictive. 

\textbf{Causal Regularizer.}
Now, given Section \ref{sec:detector}, assume that we have a classifier that outputs $c_i = \mathbb{P}[\text{$X_i$ does not cause $Y$}]$, we can design the following regularizer to encourage learning a causal predictive model:
\begin{equation}
\widehat{\mathbf{w}} = \argmin_{\mathbf{w}} \left\{\frac{1}{n} \sum_{j=1}^{n}\mathcal{L}(\xb_j, y_j|\mathbf{w}) + \lambda\sum_{i=1}^{m}c_i|w_i| \right\},
\label{eq:reg}
\end{equation}
\noindent where $\mathcal{L}(X_1, \ldots, X_n, Y|\mathbf{w})$ is the loss function for prediction of $Y$ given $X_1, \ldots, X_n$. The above regularization term is the $L_1$-norm version of the causal regularizer which will be used in our experiments. However, we can define $L_2$-norm version similarly as $\sum_{i=1}^{m}c_iw_i^2$.

The first term in Eq. (\ref{eq:reg}) is a multivariate analysis term, whereas the regularizer is constructed using a bivariate causality score of each independent variable $X_i$ and the target variable $Y$ for $i = 1, \ldots, m$. This does not create a problem because in the design of the causal regularizer we have implicitly included the other variables as hidden variables in the analysis to allow the regularizer to be used with multivariate regression. That is, the rest of the observed independent variables can be considered as hidden variables in our bivariate causality analysis which allows proper regularization.
The proposed causal regularizer is also a decomposable regularizer which makes analysis of its theoretical properties easier \citep{negahban2012}.

The interplay between causation and prediction has been studied recently, see \citep{PetBuhMei15,rojas2015causal} and the references therein. In particular, the notion of a causal regularizer was previously recognized (\citeauthor{lopezpaz2016}, \citeyear{lopezpaz2016}, Page 181; \citeauthor{lopez2016discovering}, \citeyear{lopez2016discovering}) as possible, however a specific causal regularizer has never been developed and evaluated. 
Notice that using the score of a “causal-anticausal”-only classifier without including the confounding cases, as e.g. in \citep{lopez2016discovering}, cannot properly regularize a multivariate model such as logistic regression.  Moreover, a major novelty of our proposed causal regularizer is to do joint causal variable selection (the $L_1$ regularization) and prediction, but the idea in \citep{lopez2016discovering} cannot.

\subsubsection{Analysis of Causal Regularization}
The following theorem uses a simple setting to quantify the impact of the $L_2$-norm based causal regularizer.
\begin{theorem}
\label{thm}
Consider the following general linear model:\footnote{We have intentionally made the settings of this theorem simple to have readable results. It is possible to obtain results on more general settings, potentially at the expense of cluttering the results.}
\begin{align*}
Y & = \beta_1 X_1 + \beta_2 X_2 + \varepsilon,
\end{align*}
where the noise variable $\varepsilon$ is a zero mean random variable with variance $\gamma^2<\infty$ and a distribution that satisfies the regularity conditions of Theorem 3.2 in \citep{white1982maximum}. We assume that $X_1$ causes $Y$ but $X_2$ does not and its correlation with $Y$ is due to an unobserved confounder. We have access to an imperfect causality detector with $\mathbb{P}[X_1 \text{ causes } Y] = 1-\epsilon$ and $\mathbb{P}[X_2 \text{ causes } Y] = \epsilon$, for $\epsilon \in [0, 1]$. Without loss of generality, assume that $\beta_1, \beta_2 \geq 0$. Under this setting, the causality accuracy of an estimate $\widehat{\bm{\beta}}$ is defined as follows:
\begin{equation*}
C\left(\widehat{\bm{\beta}}\right) = \mathbb{P}\left[\widehat{\beta}_{1} > \widehat{\beta}_{2}\right].
\end{equation*}
Consider the fixed design setting where an i.i.d. sample of size $n$ is drawn from the model as follows:
\begin{equation*}
\mathbf{y} = \mathbf{X}\bm{\beta} + \bm{\varepsilon},
\end{equation*}
where $\mathbf{y} \in \mathbb{R}^{n}$, $\mathbf{X} \in \mathbb{R}^{n\times 2}$, and $\bm{\varepsilon} \in \mathbb{R}^n$. For cleanness of the results, we study the orthonormal design setting where $\mathbf{X}^{\top}\mathbf{X} = nI_{2\times2}$.
Using this sample, we obtain two estimates for $\bm{\beta}$: $\widehat{\bm{\beta}}_{\text{R}, \lambda}$ and $\widehat{\bm{\beta}}_{\text{C}, \lambda}$ which are the the result of $L_2$-norm and $L_2$-norm based causally regularized regression, respectively.  Asymptotically, as $n\to\infty$, we have the following results:
\small
\begin{align}
C\left(\widehat{\bm{\beta}}_{\text{C}, \lambda}\right)& = \Phi\left(\frac{\sqrt{n}}{\gamma}\frac{(1+\lambda(1-\epsilon) )\beta_1-(1+\lambda \epsilon)\beta_2}{\sqrt{(1+\lambda \epsilon)^2+ (1+\lambda (1-\epsilon))^2}}\right), \label{eq:res_cause}\\
C\left(\widehat{\bm{\beta}}_{\text{R}, \lambda}\right) & = \Phi\left(\frac{\sqrt{n}}{\gamma}\frac{(\beta_1 - \beta_2)}{\sqrt{2}}\right), \label{eq:res_l2}
\end{align}
\normalsize
where $\Phi(\cdot)$ denotes the CDF of the unit Gaussian distribution.
\end{theorem}
A proof is provided in Appendix \ref{sec:proof}. To understand the result, considering several special cases can be helpful. When the causal detector is perfect ($\epsilon=0$), we can rewrite $C\left(\widehat{\bm{\beta}}_{\text{C}, \lambda}\right)$ as follows
\begin{equation*}
C\left(\widehat{\bm{\beta}}_{\text{C}, \lambda}\right) = \Phi\left(\frac{\sqrt{n}}{\gamma}\frac{(1+\lambda )\beta_1-\beta_2}{\sqrt{1+ (1+\lambda )^2}}\right).
\end{equation*}
Compared to Eq. (\ref{eq:res_l2}), we see a $1+\lambda$ factor scaling of the causal coefficient $\beta_1$ against the non-causal coefficient $\beta_2$ in the nominator, increasing the chance of correct causality detection. That is, a perfect causality detector guarantees causal interpretability if the magnitude of $\lambda$ outweights the predictive advantage of $\beta_2$ over $\beta_1$. 
When the causal detector is random ($\epsilon=1/2$), we can show that $C\left(\widehat{\bm{\beta}}_{\text{C}, \lambda}\right) = C\left(\widehat{\bm{\beta}}_{\text{R}, \lambda}\right)$. That is, a non-informative causality detector makes causal regularization equivalent to standard $L_2$ regularization. Finally, in the limit of large penalization coefficient, we obtain:
\begin{equation*}
\lim_{\lambda\to \infty}C\left(\widehat{\bm{\beta}}_{\text{C}, \lambda}\right) = \Phi\left(\frac{\sqrt{n}}{\gamma}\frac{(1-\epsilon )\beta_1-\epsilon\beta_2}{\sqrt{\epsilon^2+ (1-\epsilon )^2}}\right).
\end{equation*}
The impact of the error rate of the causality detector in the nominator can be seen as linear scaling of the causal coefficient by $(1-\epsilon)$ and the non-causal factor by $\epsilon$. 


Another property of the causal regularizer is that the two-step analysis can be cast as a form of causal regularization where we use hard scores instead of soft scores. Consider the following setting:
\begin{equation*}
\widehat{\mathbf{w}} = \argmin_{\mathbf{w}} \left\{ \frac{1}{n} \sum_{j=1}^{n}\mathcal{L}(\xb_j, y_j|\mathbf{w}) + \gamma\sum_{i=1}^{m}c'_i|w_i| \right\},
\end{equation*}
where $c'_i =  \varepsilon$ if $c_i \leq 1/2$ and $1-\varepsilon$ otherwise.
Now, consider the limiting case of $\varepsilon\to 0$ and $\gamma\varepsilon \to \lambda$. This case corresponds to the two-step procedure with $L_1$ regularized logistic regression.

\subsection{Causal Regularizers in Neural Networks}
\label{sec:hyp}
\begin{figure}[t]
    \centering
    \begin{subfigure}[t]{0.4\textwidth}
    \centering
        \includegraphics[scale=0.45]{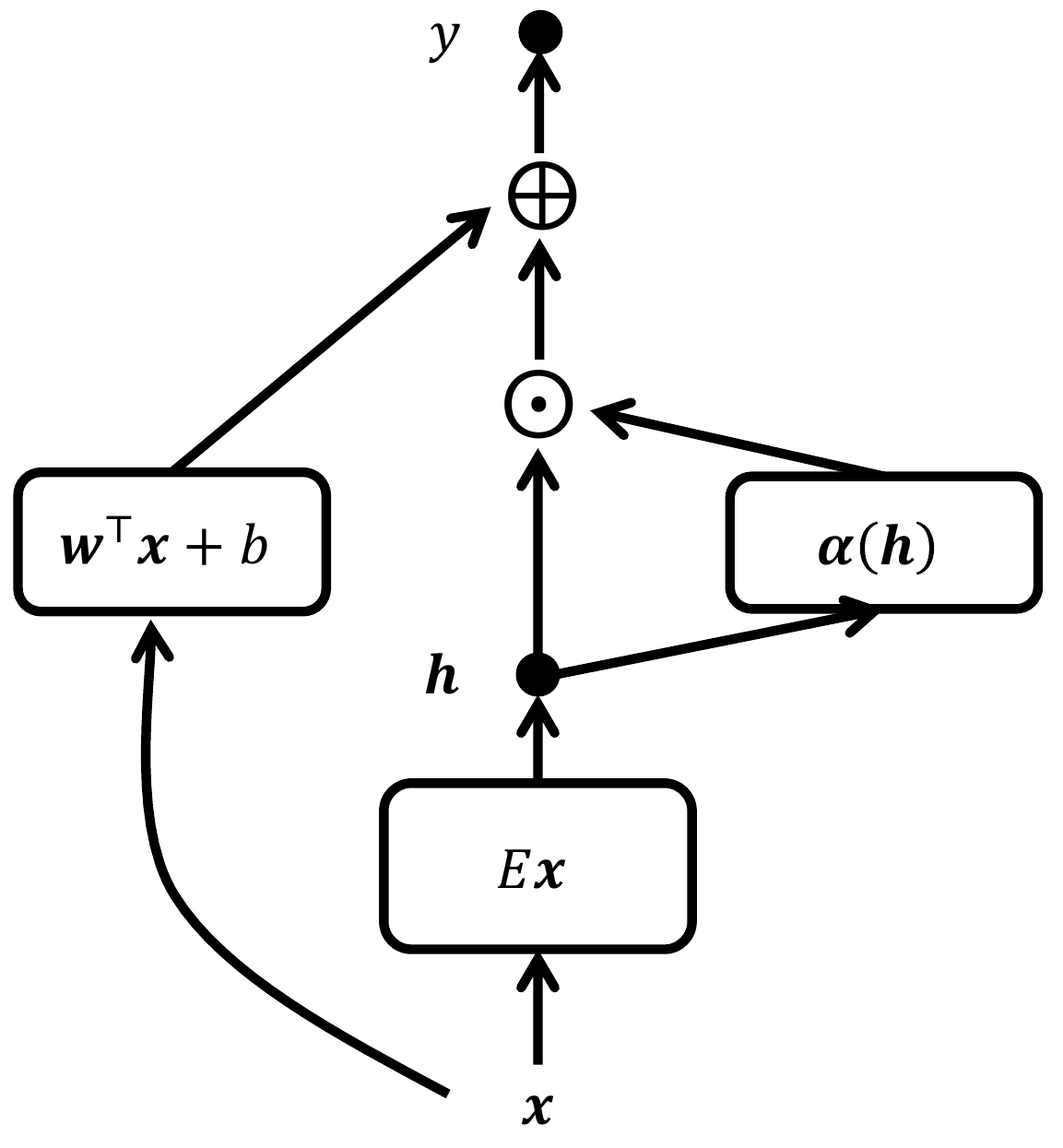}
        \caption{Non-linear causality analysis}
        \label{fig:nonlin}
    \end{subfigure}
    \quad
    \begin{subfigure}[t]{0.4\textwidth}
    \centering
        \includegraphics[scale=0.45]{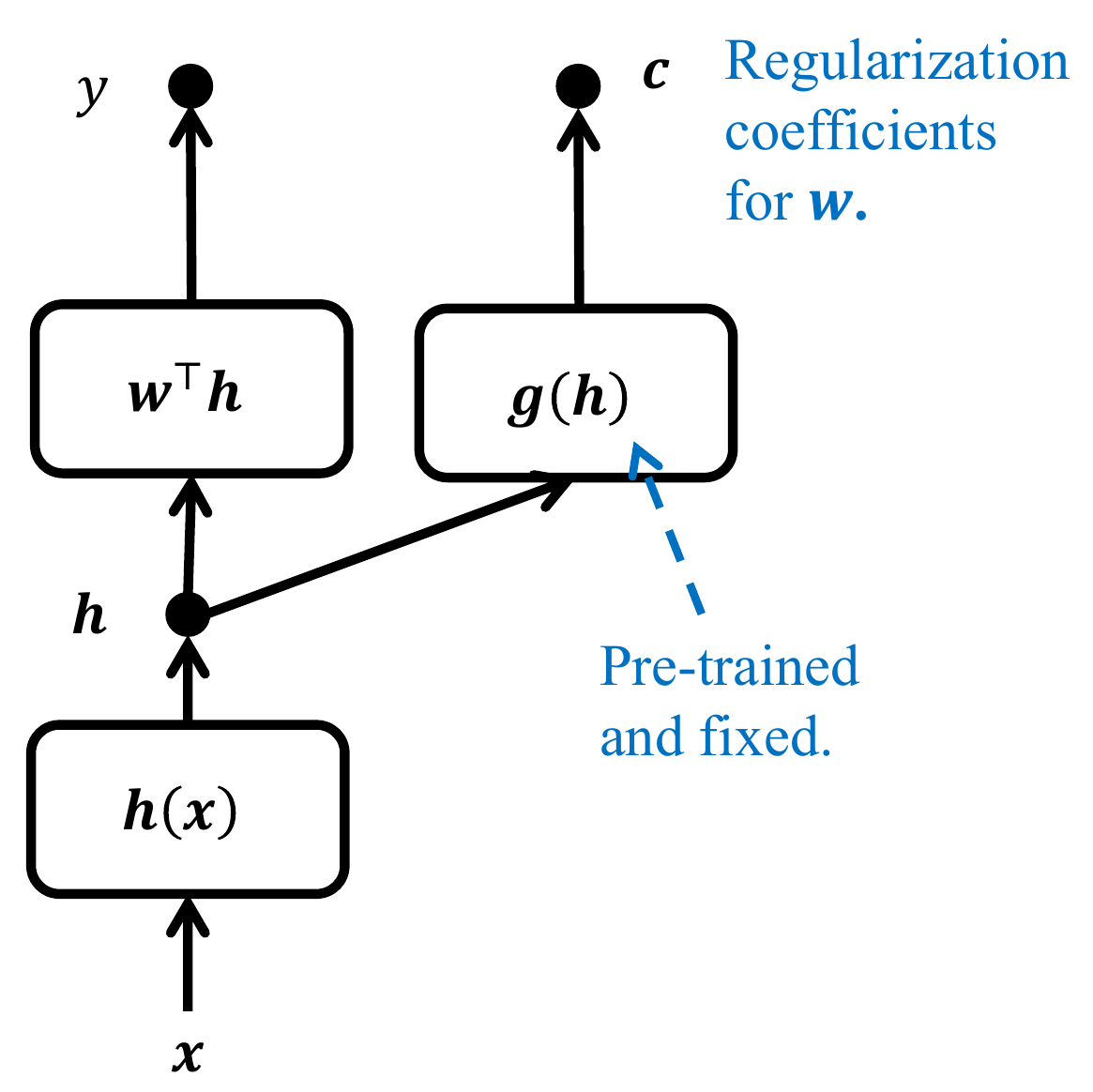}
        \caption{Multivariate causal hypothesis generation}
        \label{fig:hype_gen}
    \end{subfigure}
    \caption{Two scenarios of using the proposed causal regularizer: 
    (\subref{fig:nonlin}) In the proposed architecture, applying the causal regularizer allows identification of causal relationships in the non-linear settings, where the causality coefficient can change from subject to subject. (\subref{fig:hype_gen}) The causal regularizer allows us to explore the high-dimensional multivariate combinations of the variables and identify plausible hypotheses. Here, $\bm{g}$ generates the causal regularization coefficients for the hypotheses $\bm{h}$. The regularizer encourages the coordinates of $\bm{h}$ to be more causal.}\label{fig:usecases}
    \vspace{-0.1in}
\end{figure}

We demonstrate two key scenarios of using the causal regularizer as shown in Figure \ref{fig:usecases}. 

\textbf{Non-linear Modeling.}
The linear model in Eq. (\ref{eq:reg}) assumes that the strength of the impact of each independent variable on the target variable is fixed. However, according to probabilistic view of causation \citep{pearl2009causality}, the strength of causation can change from subject to subject. Thus, we need non-linear extensions of logistic regression that can be regularizerd by the causal regularizer and steered towards being causal.  

To address this problem, we seek neural network architectures that represent the impact of each independent variable by a single coefficient (that can change for each subject) and regularize the coefficients with the causal regularizer. In particular, we propose  the following non-linear generalized linear model:
\begin{equation}
\sigma^{-1}(\mathbb{E}[Y|\mathbf{x}]) =\mathbf{w}^{\top}\mathbf{x} + \bm{\beta}^{\top}(\bm{\alpha}(E\mathbf{x})\odot(E\mathbf{x})) + b,
\label{eq:nonlin}
\end{equation}
\noindent where the embedding matrix $E\in \mathbb{R}^{q\times m}$ maps the input $\mathbf{x}\in \mathbb{R}^{m}$ to a lower dimensional representation space and the symbol $\odot$ denotes the element-wise product. 
The logistic sigmoid function $\sigma$ maps the real values to the $[0, 1]$ interval. The term $\mathbf{w}^{\top}\mathbf{x}$ acts as the skip connection and is initialized by the result of the logistic regression. The embedding allows dealing with very large set of discrete concepts and can be initialized via techniques such as skip-gram \citep{mikolov2013distributed} or GloVe \citep{pennington2014glove}. The vector $\bm{\alpha}(E\mathbf{x})$ is computed using a Multilayer Perceptron (MLP).  

The model in Eq. (\ref{eq:nonlin}) is a non-linear extension of logistic regression that is suitable for causal regularization. We can reorder the equations to write the right hand side of Eq.~(\ref{eq:nonlin}) as $\bm{\omega}(\mathbf{x})^{\top}\mathbf{x} + b$, where the new regression coefficient $\bm{\omega}$ can change with every input. Each coordinate of the new regression coefficient can be calculated as $\omega_i(\mathbf{x}) = w_i + (\bm{\beta}\odot\bm{\alpha}(E\mathbf{x}))^{\top}E_i$, where $E_i$ denotes the $i$th column of the embedding matrix $E$. The variability of $\omega_i(\mathbf{x})$ for each input $\mathbf{x}$ enables us to perform individual causality analysis. For training, we can penalize the $\bm{\omega}$ coefficients and  minimize the following loss function
\begin{equation}
\frac{1}{n}\sum_{j=1}^{n}\left\{ \widetilde{\mathcal{L}}(\mathbf{x}_j, y_j) + \lambda\sum_{i=1}^{m}c_i\omega_i^2(\mathbf{x}_j)~\right\},
\end{equation}
where $\widetilde{\mathcal{L}}$ denotes the negative log-likelihood of the model described by Eq. (\ref{eq:nonlin}). The change of the prediction vector with each sample $\mathbf{x}$ can be related to the probabilistic definition of causation \citep{pearl2009causality} in the sense that the strength of causality may change from a subject to another one.

\textbf{Multivariate Causal Hypothesis Generation.}
A key application of our proposed causal regularizer in conjunction with deep representation learning is to efficiently extract multivariate causal hypotheses from the data.
Figure \ref{fig:hype_gen} shows an example of causal hypothesis generation where the hypotheses are generated via an MLP.
We assume that there is a representation learning network with $K$-dimensional output $\bm{h}(\mathbf{x}) \in \mathcal{I}^{K}$, where $\mathcal{I}$ denotes the range of the output, for example $\mathcal{I} = (0, 1)$ for sigmoid and $\mathcal{I} = [0, \infty)$ for ReLU activation functions. Our goal is to force each dimension of $\bm{h}$ to be causal, thus each coordinate of $\bm{h}(\mathbf{x})$ can be used as a multivariate causal hypothesis. In particular, we aim at minimizing the following objective function:
\begin{equation}
\frac{1}{n}\sum_{j=1}^{n}\left\{\mathcal{L}(\mathbf{w}^{\top}\bm{h}_j+b) + \lambda \sum_{i=1}^{K}g_i(\bm{h}_{j})|w_i|\right\}.
\label{eq:hypo}
\end{equation}
Our approach is to train an anti-causality detector based on \citep{lopez2016discovering} and design the regularizer $\bm{g}(\bm{h}(\mathbf{x}))$ based on its score. Then, as shown in Figure \ref{fig:hype_gen}, we can combine it with the neural network to regularize the coefficients of the last layer of the MLP which predicts the labels from $\bm{h}$. The weights of the lower layers in $\bm{h}(\mathbf{x})$ are regularized using $L_1$ regularizer to make the generated causal hypotheses simple and interpretable. 

The learning process has two steps: First, the causality detector network $\bm{g}(\bm{h})$ is trained on a synthetic dataset with causal and anti-causal scenarios are labeled as $\ell=0$ and $\ell=1$, respectively. We select the non-linearity for $\bm{h}$ to be the logistic sigmoid function, thus we use Beta distribution for generating synthetic data for training of the causality classifier. In the second phase, the coefficients of $\bm{g}(\bm{h})$ are fixed and we train the rest of the parameters in Eq. (\ref{eq:hypo}). 
To train the network, we select batches with fixed-size of 200 examples. The size of the batches indicate the number of samples from $\bm{h}$ that is available to the causality detector. We select this number to be large enough such that error rate of the causality detector in \citep{lopez2016discovering} becomes lower than $2\%$. 


\section{Experiments}
\label{sec:exp}
We evaluate the proposed causal regularizer in Section \ref{sec:argue} both in terms of its predictive and causal performance.  Next, we compare the quality of the codes identified as causes of heart failure identified by different approaches. Finally, we evaluate performance of multivariate causal hypothesis generation by qualitatively analyzing the extracted hypotheses. We defer evaluation of the causality detection algorithms to Appendix \ref{sec:details_detector}, as they are not the main contributions of this work. Table \ref{tab:sybmols} lists the acronyms and symbols for techniques used in the experiments to improve the presentation.

\begin{table}[t]
\centering
\small
\caption{List of symbols and acronyms used in the experiment section. \textbf{Bold font} shows our proposed approaches. }
\label{tab:sybmols}
\begin{tabular}{@{}l|p{6cm}||l|p{6cm}@{}}
Symbol & Description & Symbol & Description \\
\midrule
CD & Causality detector, described in Section \ref{sec:detector} & $c_{\text{CD}}$ & The output of a causality detector\\
\textbf{LogCause} & Logistic regression regularizerd by the causal regularizer & LogL$_1$ & Logistic regression regularizerd by the $L_1$ regularizer  \\
Two-step & The two step procedure of causal variable selection and $L_1$ logistic regression, as discussed in Section \ref{sec:argue} & $\mathbf{w}_{\mathrm{Algo}}$ & The regression coefficients of an algorithm (one of LogCause, Two-step, or LogL$_1$) \\
\textbf{nonlinCause} & The non-linear causality analysis model in Eq. (\ref{eq:nonlin}) & \textbf{CauseHyp} & The multivariate causal hypothesis generation described in Eq. (\ref{eq:hypo}) \\
\bottomrule
\end{tabular}
\normalsize
\end{table}

\begin{table*}[t]
\centering
\caption{Prediction accuracy results on two datasets. (mean$\pm$standard deviation)}
\label{tab:pred_results}
\begin{tabular}{l|c|c|c|c}
\multirow{ 2}{*}{Algorithms}& \multicolumn{2}{|c|}{Heart Failure} & \multicolumn{2}{|c}{MIMIC III}\\
& AUC & $F_1$ & AUC & $F_1$\\
\midrule
LogCause & $0.8289\pm 0.0064$ &$0.4147\pm 0.0192$ & $0.9772\pm 0.0022$ & $0.7871\pm 0.0097$ \\
LogL$_1$& $0.8289\pm 0.0054$&$0.4109\pm 0.0150$ & $0.9774\pm 0.0022$ & $0.7869\pm 0.0095$\\
Two-Step &$0.7276\pm 0.0086$ &$0.2686\pm 0.0134$ & $0.9515\pm 0.0033$ & $0.6745\pm 0.0106$\\
\bottomrule
\end{tabular}
\end{table*}

\begin{figure*}[t]
    \centering
    \begin{subfigure}[t]{0.16\textwidth}
    \centering
        \includegraphics[width=\textwidth]{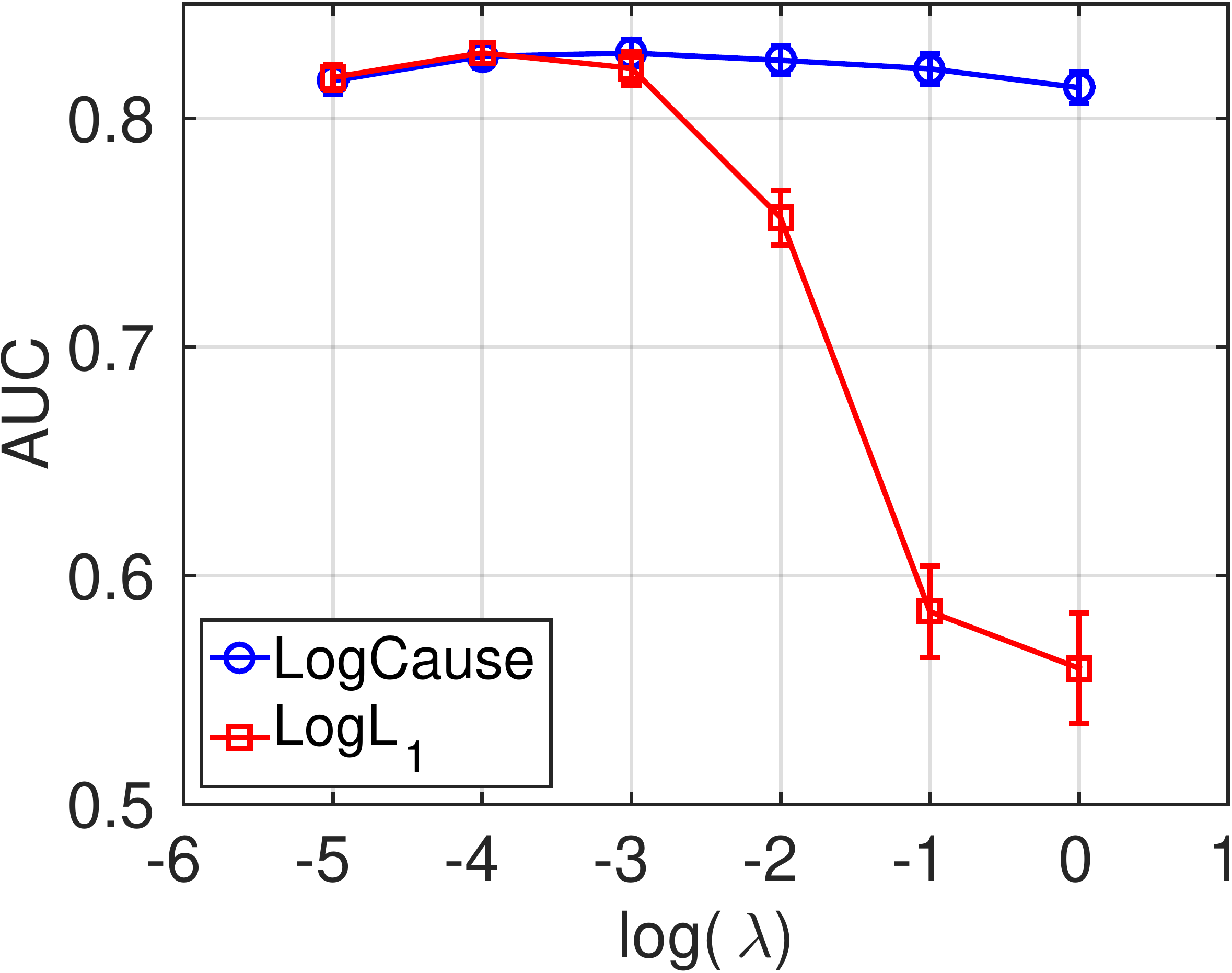}
        \caption{AUC on HF}
        \label{fig:auc_sutter}
    \end{subfigure}
    \begin{subfigure}[t]{0.16\textwidth}
    \centering
        \includegraphics[width=\textwidth]{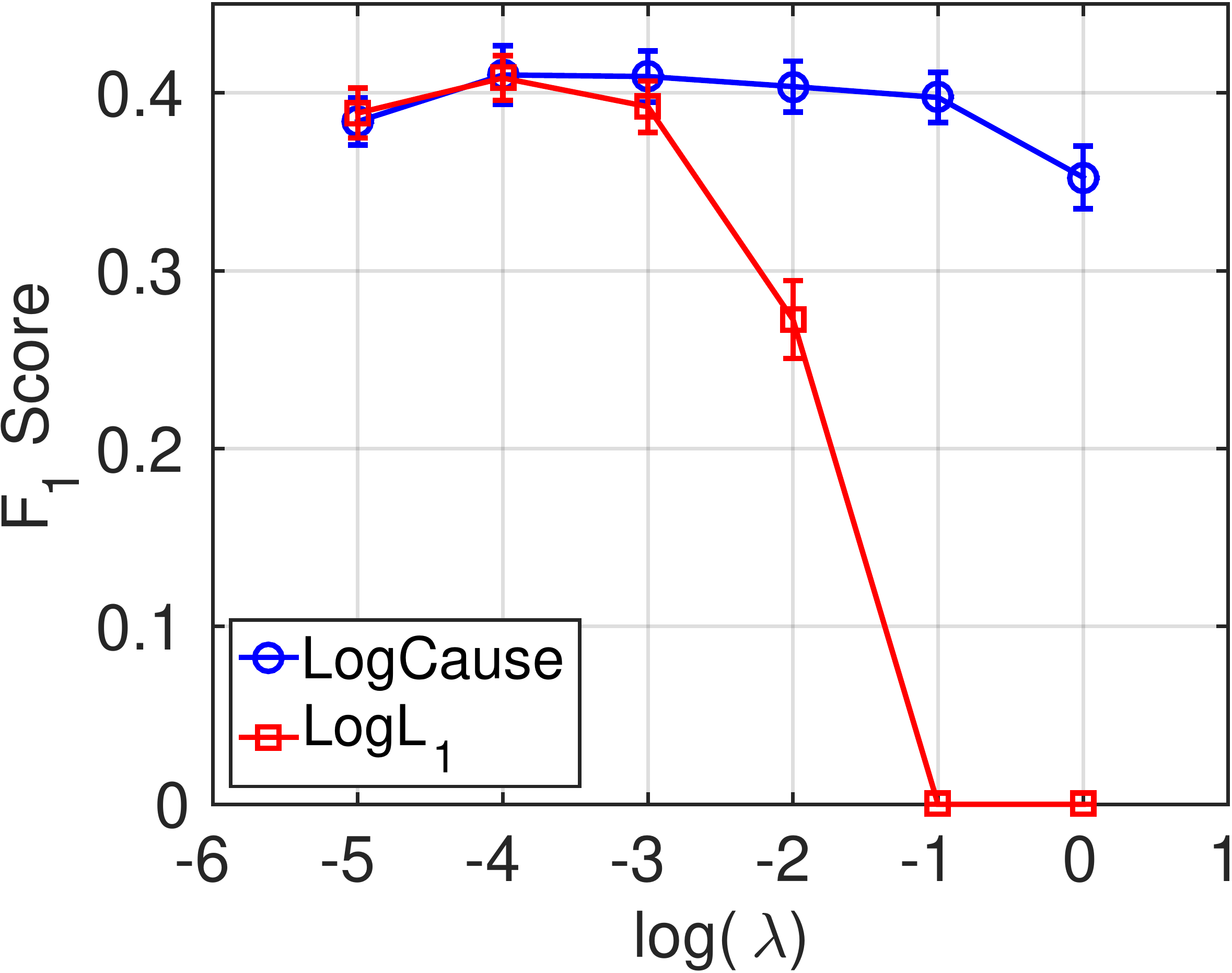}
        \caption{$F_1$ on HF}
        \label{fig:f1_sutter}
    \end{subfigure}
    \begin{subfigure}[t]{0.16\textwidth}
    \centering
        \includegraphics[width=\textwidth]{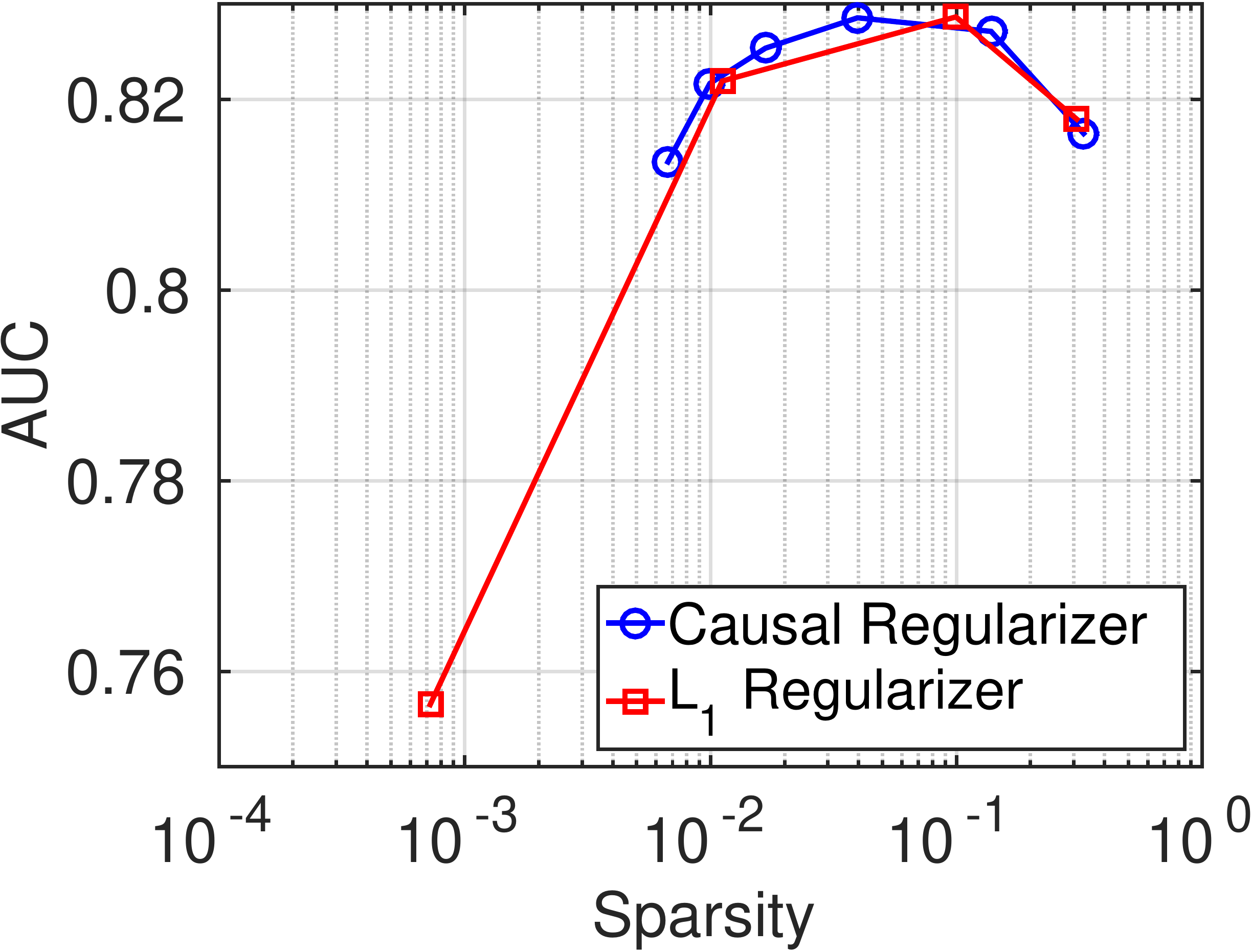}
        \caption{\scriptsize AUC v. Sparsity (HF) \normalsize}
        \label{fig:sp_auc_sutter}
    \end{subfigure}
    \begin{subfigure}[t]{0.16\textwidth}
    \centering
        \includegraphics[width=\textwidth]{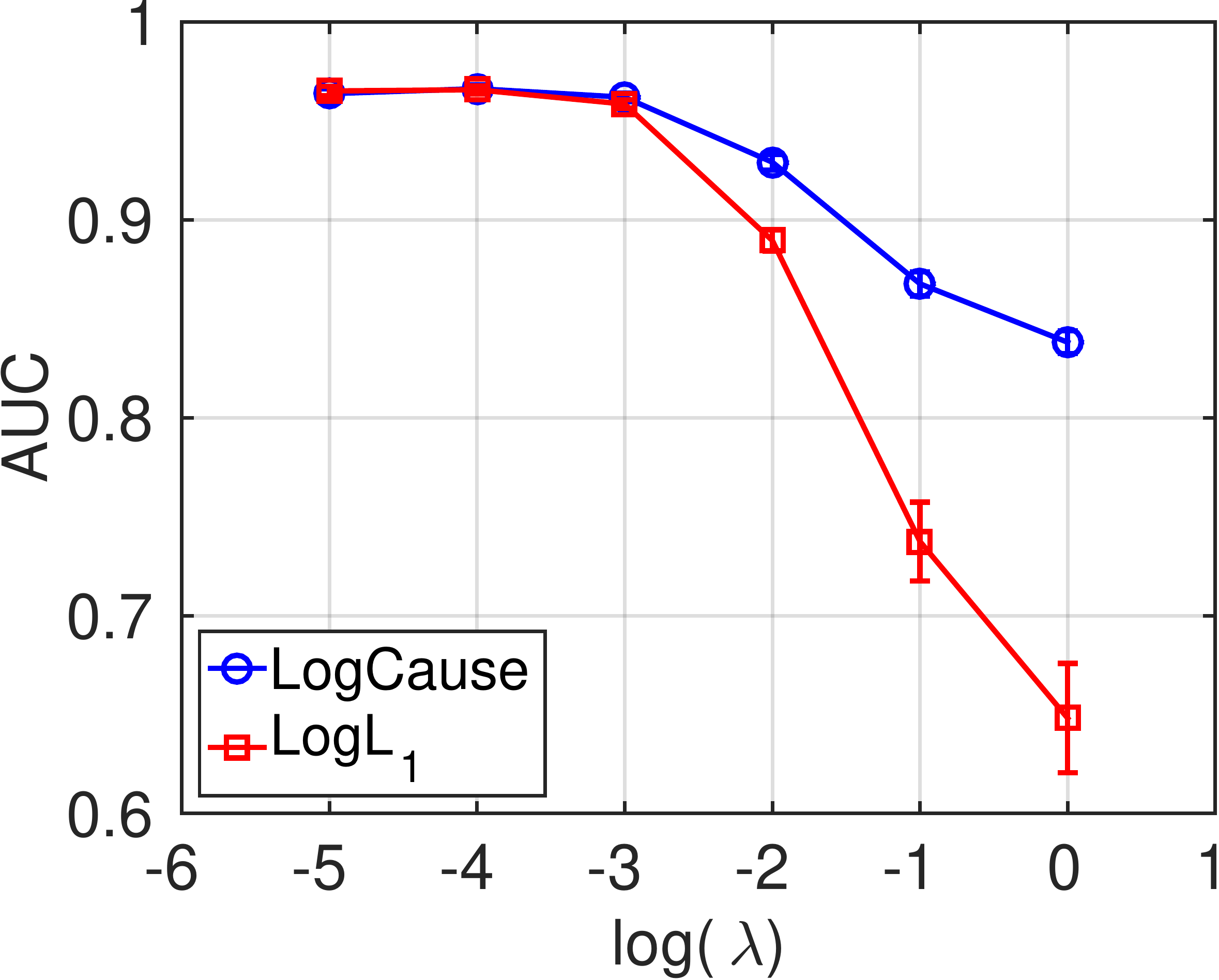}
        \caption{AUC on MIMIC}
        \label{fig:auc_mimic}
    \end{subfigure}
    \begin{subfigure}[t]{0.16\textwidth}
    \centering
        \includegraphics[width=\textwidth]{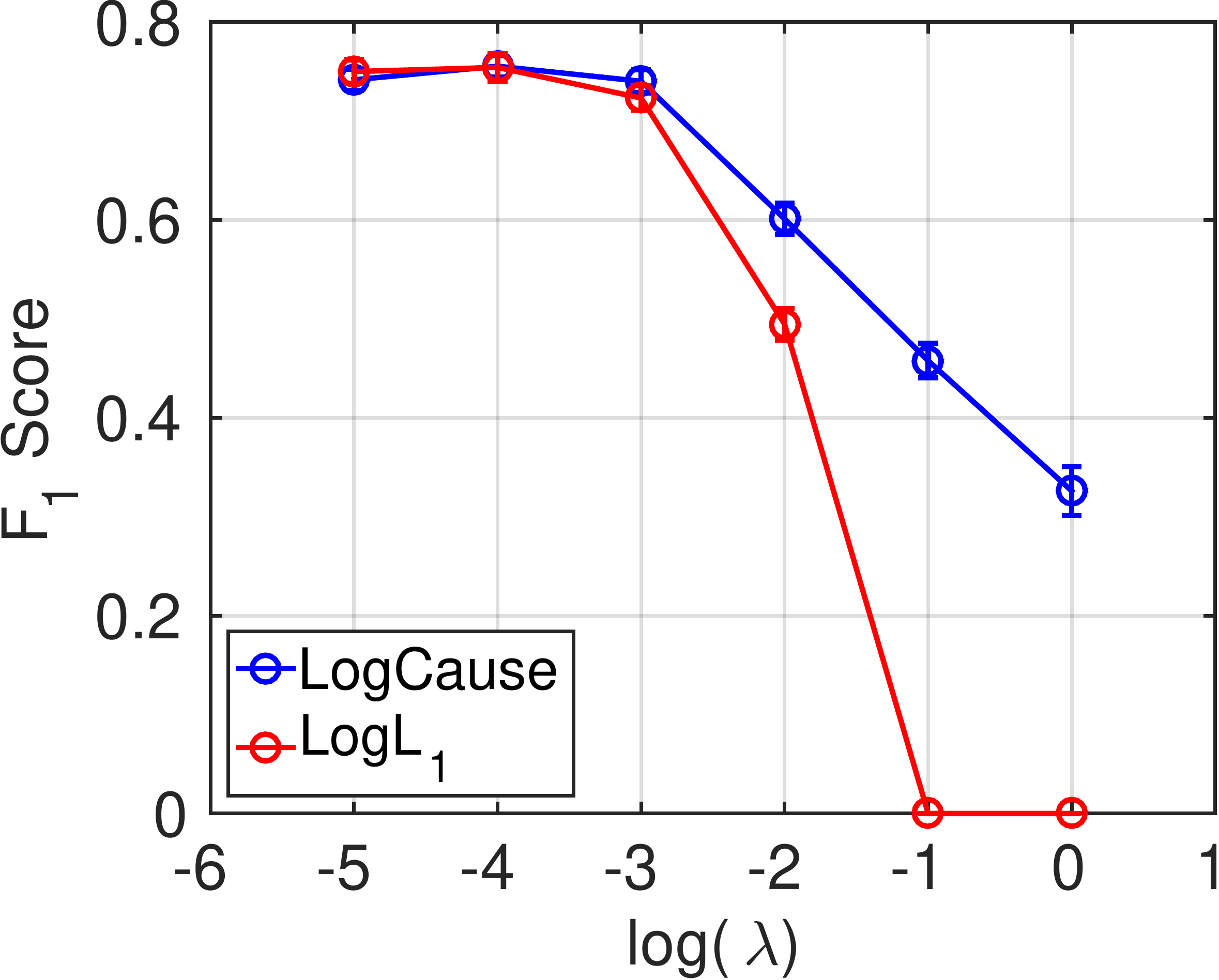}
        \caption{$F_1$ on MIMIC}
        \label{fig:f1_mimic}
    \end{subfigure}
    \begin{subfigure}[t]{0.16\textwidth}
    \centering
        \includegraphics[width=\textwidth]{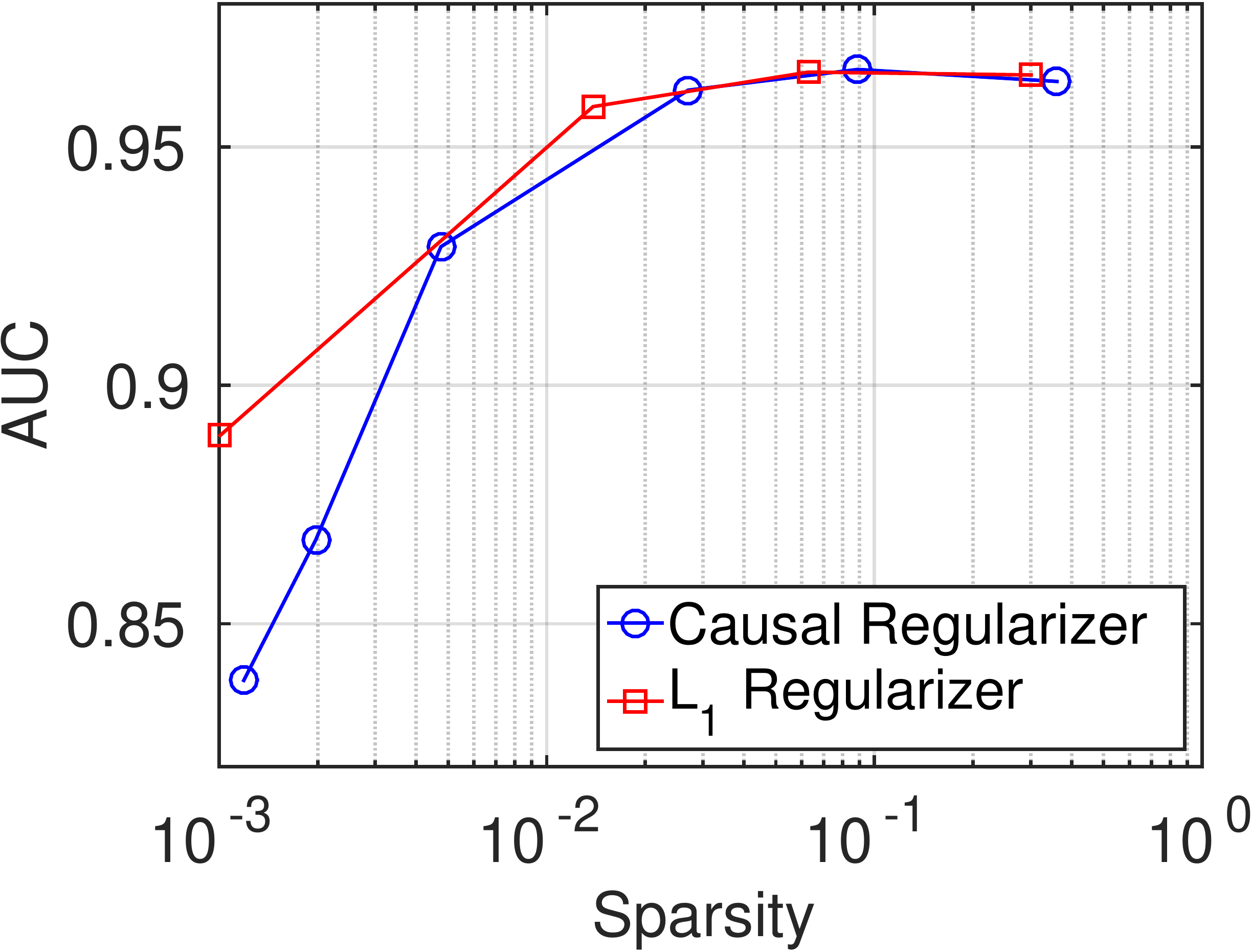}
        \caption{\tiny AUC v. Sparsity (MIMIC) \normalsize}
        \label{fig:sp_auc_mimic}
    \end{subfigure}
    \caption{Comparison of variable selection in logistic regression via the causal and $L_1$ regularizers on two datsets and two accuracy measures. Note the stability of variable selection by LogCause as the penalization coefficient varies.}\label{fig:impact_lambda}
\end{figure*}
\subsection{Data}
The \textbf{Sutter Health heart failure (HF) dataset} consists of Electronic Health Records of middle-aged adults collected by Sutter Health for study of heart failure. From the encounter records, medication orders, procedure orders and problem lists, we extracted visit records consisting of diagnosis, medication and procedure codes. We denote the set of such codes by $\mathcal{C}$. 

Given a visit sequence $\vb_1, \ldots, \vb_T$, we try to predict if the patient will be diagnosed with heart failure (HF) and identify the key causes of increase heart failure risk.  To this end, 3,884 cases are selected and approximately 10 controls are selected for each case (28,903 controls). The case/control selection criteria are fully described in Appendix \ref{sec:cohort}. Cases have index dates to denote the date they are diagnosed with HF. Controls have the same index dates as their corresponding cases. We extract diagnosis codes, medication codes and procedure codes from the 18-month window before the index date. There are in total 17,081 number of unique medical codes in this dataset.

The \textbf{MIMIC III} dataset \citep{johnson2016mimic} is a publicly available dataset consisting of medical records of intensive care unit (ICU) patients over 11 years. We use a public query\footnote{\url{https://github.com/MIT-LCP/mimic-code/blob/master/concepts/cookbook/mortality.sql}} to extract the binary mortality labels for the patients. Our goal is to use the codes in the patients' last visit to the ICU and predict their mortality outcome. Our dataset includes 46,520 patients out of whom 5810 have deceased (mortality=1). A totoal of 14,587 different medical codes are used in this dataset.


\textbf{Feature construction.} Given the sequence of visits $\vb_1^{(i)}, \ldots, \vb_T^{(i)}$ for patients $i = 1, \ldots, n$, we create a feature vector $\xb_i \in \mathbb{N}_0^{|\mathcal{C}|}$ by counting the number of codes observed in the records of the $i$th patient. Given the large variations in the number of codes, we logarithmically bin the count data into 16 bins. The final data is in the form of $(\xb_i, y_i)$ where $y_i$ is $i$th patient's label; heart failure and mortality outcome in the heart failure and MIMIC III datasets, respectively.

\textbf{Training details.} Because we generate synthetic datasets for training the causality detector neural networks, we can generate as many new batches of data for training and parameter tuning purposes as required. 
For training and parameter tuning of the models in Section \ref{sec:method}, we perform the common 75\%/10\%/15\% training/validation/test splits. The full details of the training procedure for the neural networks are given in Appendix \ref{sec:neural_cause}.

\begin{figure*}[t]
    \centering
    \begin{subfigure}[t]{0.3\textwidth}
    \centering
        \includegraphics[width=\textwidth]{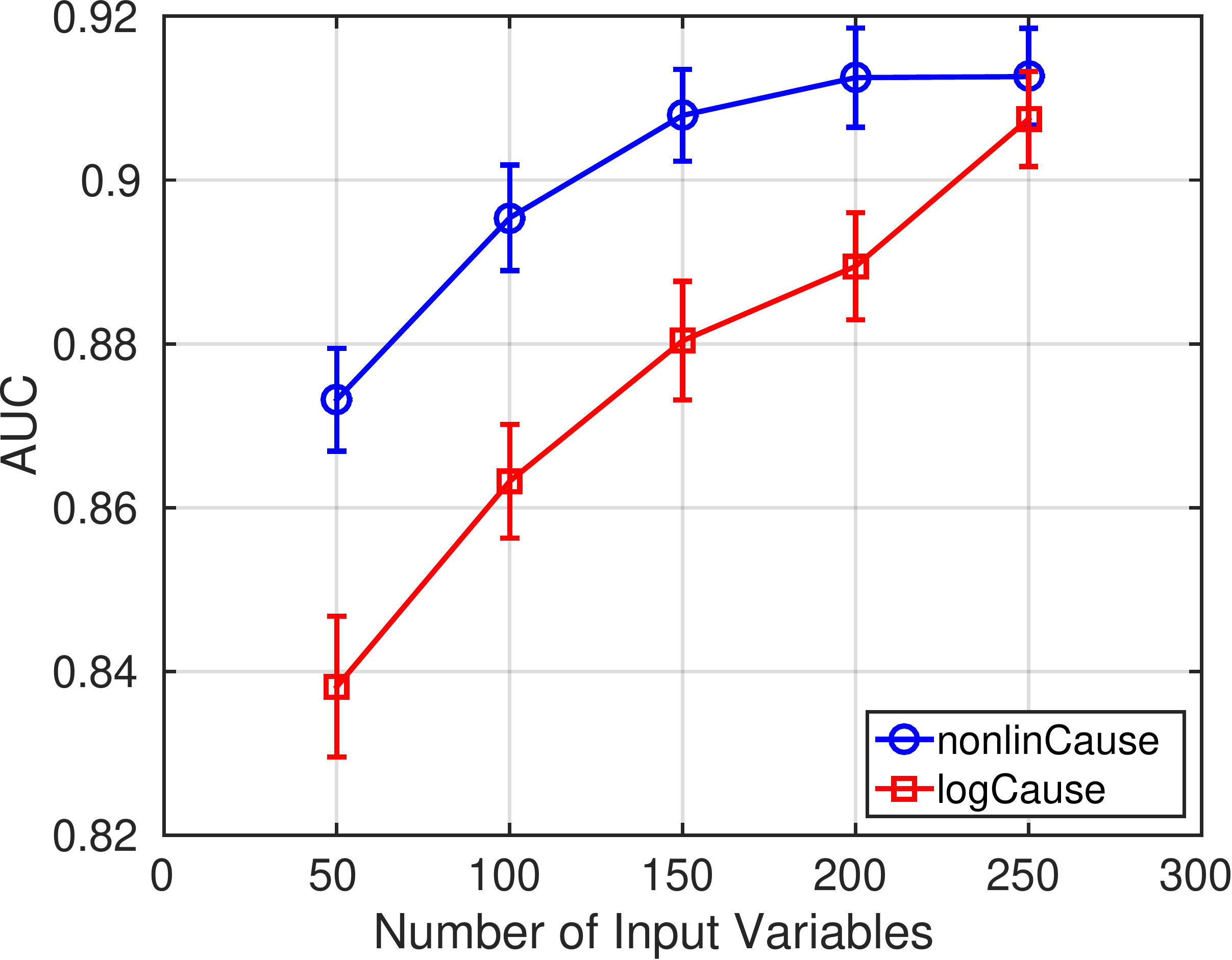}
        \caption{Predictive gain by nonlinCause}
        \label{fig:nonlin_gain}
    \end{subfigure}
    \quad
    \begin{subfigure}[t]{0.3\textwidth}
    \centering
        \includegraphics[width=\textwidth]{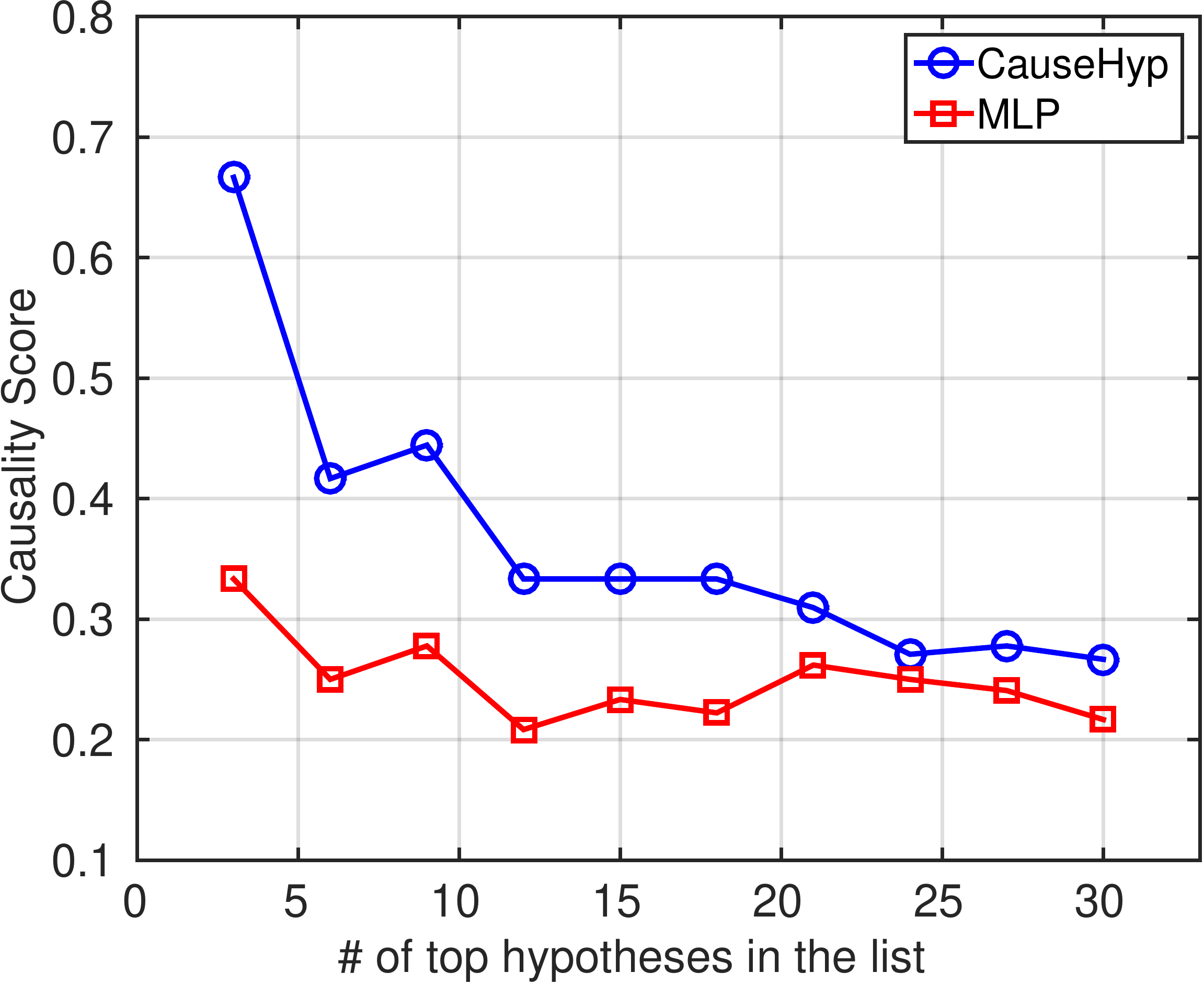}
        \caption{Accuracy of CauseHyp}
        \label{fig:hypo_gen}
    \end{subfigure}
    \quad
    \begin{subfigure}[t]{0.3\textwidth}
    \centering
        \centering
    \includegraphics[width=\textwidth]{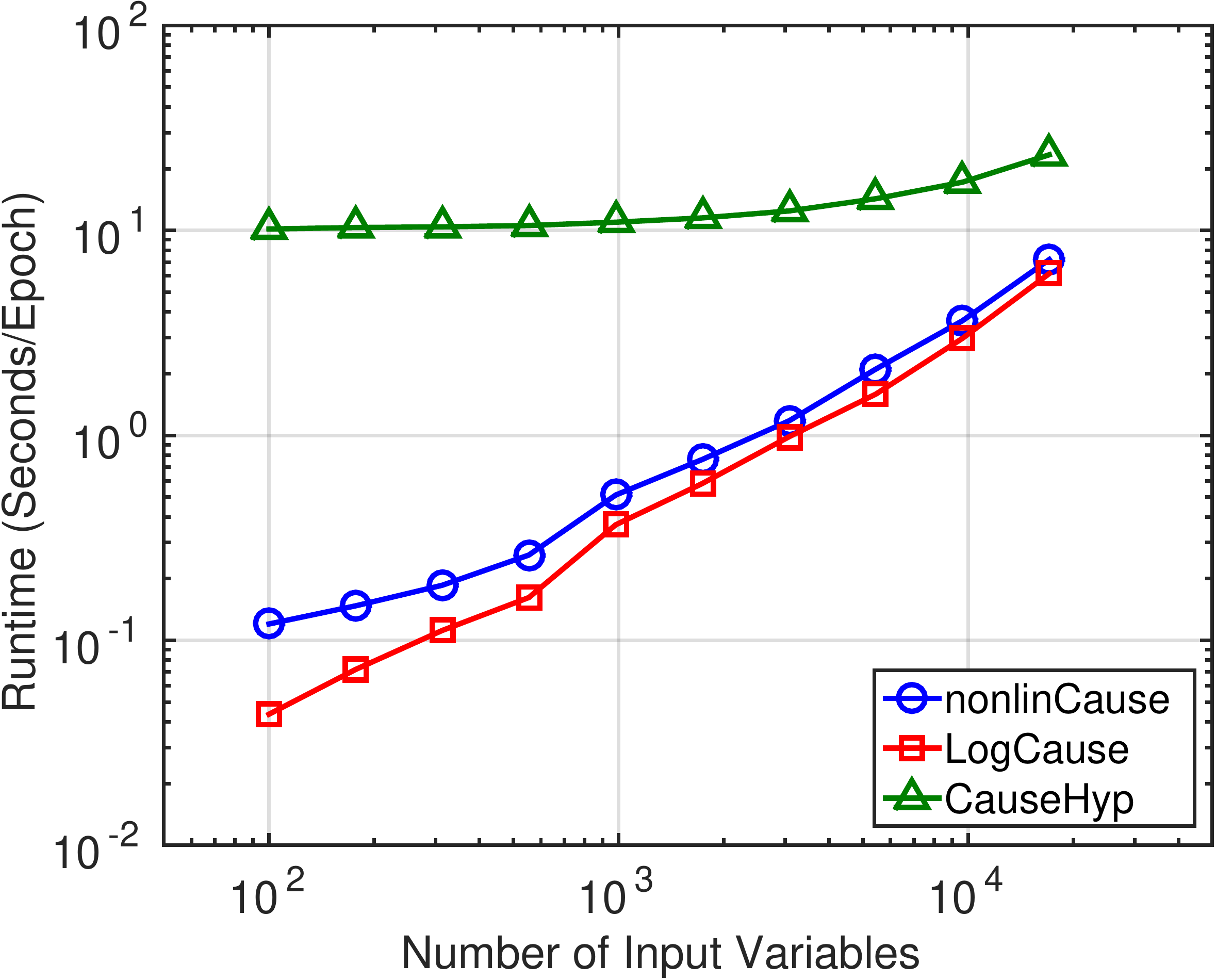}
    \caption{Runtime}
    \label{fig:scalability}
    \end{subfigure}
    \caption{(\subref{fig:nonlin_gain}) The predictive gain by nonlinCause on the MIMIC III datset.  The gain is more visible when fewer features are used in the analysis because the input become more expressive by themselves. We select the variables in the descending order of variance. (\subref{fig:hypo_gen}) Average causality score computed using ground truth causality labels for generated hypotheses. We compute the score for top $k$ hypotheses reported by two algorithms. (\subref{fig:scalability}) Runtime of the proposed algorithms as number of input variables change.}
\end{figure*}

\renewcommand{\arraystretch}{0.98}
\begin{table*}[h!]
\centering
\footnotesize
\caption{Examples of multivariate causal hypotheses generated via causal regularizer.}
\label{tab:hypo}
\begin{tabular}{@{}l@{\;\,}|@{\;\,}l@{\;\,}|@{\;\,}l@{}}
\textbf{Name} & \textbf{Conditions} & \textbf{Description}\\
\midrule
\midrule
\multirow{5}{*}{\parbox{2cm}{Aortic Dissection from Trauma}} & Dissection of aorta & \multirow{5}{*}{\parbox{8.5cm}{This collection of diagnoses is is especially causal for heart failure, as heart failure can manifest as a complication of dissection of aorta. Dissection of aorta can present with abdominal pain, and may happen in traumatic injuries that involve burn of unspecified degree of other and multiple sites of trunk, occurring together. }}\\
& Burn in multiple sites of trunk & \\
& Abdominal pain, lower left quadrant & \\
& & \\
& & \\
\midrule
\multirow{4}{*}{\parbox{2cm}{Kidney Neoplasm and Severe Infections}} & Malignant neoplasm of kidney & \multirow{4}{*}{\parbox{8.5cm}{Neoplasms in the kidney may lead to paraneoplastic systemic effects that may lead to heart failure. Furthermore, having concurrent severe infections such as tuberculosis can also increase the risk of heart failure.
}}\\
& History of infectious and parasitic diseases & \\
& Tuberculosis of lung & \\
& & \\
\midrule
\multirow{5}{*}{\parbox{2cm}{Metabolic Syndrome with Concurrent Infections and Pregnancy}} & Metabolic syndrome & \multirow{5}{*}{\parbox{8.5cm}{Metabolic syndrome co-occurring with severe infections such as tuberculosis can lead to heart failure. Obstetrical pulmonary embolisms can lead to acute heart failure. }}\\
& Tuberculosis of lung& \\
& Obstetrical pulmonary embolism & \\
& & \\
& & \\
\bottomrule
\end{tabular}
\end{table*}

\subsection{Predictive performance evaluation}
\label{sec:reg_eval}
Table \ref{tab:pred_results} shows the test accuracy of heart failure and mortality prediction in heart failure and MIMIC datasets, respectively. We have run each algorithm ten times and report the mean and standard deviation of the performance measures.  As we can see, the proposed causal regularizer does not hurt the predictive performance, whereas the two-step procedure significantly reduces the accuracy.

An interesting phenomenon, shown in Figure \ref{fig:impact_lambda}, is the relative robustness of the performance with respect to the value of the penalization parameter compared to the $L_1$ regularization case. This robustness comes at no surprise, because the causal regularizer assigns very small penalization coefficients to the causal variables and as we discussed in Section \ref{sec:argue}, only with very high values of penalization we can force all coefficients to become zero, see Figures \ref{fig:sp_auc_sutter} and \ref{fig:sp_auc_mimic} which show the  sparsity results. The predictive robustness of the causal regularizer can be also partially attributed to the invariant prediction \cite{PetBuhMei15} property of causal models. That is, the robustness can be due to the fact that the causal regularizer might match the true generative process of the dataset better than the flat $L_1$ regularizer and put the model under less pressure as we increase the penalization parameter. We demonstrate the predictive gain by nonlinCause in Figure \ref{fig:nonlin_gain}. Furthermore, the impact of changing the regularization parameter on the number of selected variables is visualized in Figures \ref{fig:mi-cause} and \ref{fig:impact_lamda_vis} in Appendix \ref{sec:cause_eval}.

\subsection{Causality detection performance evaluation}
The risk factors for heart failure are well-studied in medical literature, making the heart failure condition an ideal case for study of causality.
To evaluate the causality detection performance of the algorithms, we generate top 100 influential factors by each method. We ask a clinical expert to label each factor as ``causal'', ``not-causal'', and ``potentially causal'' and assign scores $1$, $0$, and $0.5$ to them, respectively. To prevent bias by the expert, we ask him to label a single list of all unique codes in the three lists and use this list to find the scores for individual lists.
Figure \ref{fig:causality_intro} shows the average causality score by each algorithm based on the labels provided by the medical expert.  As expected, $L_1$ regularized logistic regression performs poorly, as it is susceptible to the impact of confounded variables. Performance of the causally regularized logistic regression is superior to the two step procedure, which suggests that picking factors that are both causal and highly predictive leads to better causality score. The result in Figure \ref{fig:causality_intro} together with the predictive results in Table \ref{tab:pred_results} confirm that the causal regularizer can be efficiently used for finding few causal variables that are highly predictive of the target quantity.

The qualitative advantages of the regularized approach can also be seen by the results in Table \ref{tab:regu_adv}  in Appendix \ref{sec:tables}.  We have marked the disease codes that can potentially increase the risk of heart failure, but the \textit{predicted} causality score $c_{\text{CD}}$ for them is lower than $0.5$ and the two-step procedure would have eliminated from the predictors set (as shown in Table \ref{tab:mi_list} in Appendix \ref{sec:tables}). Thus, the causal regularizer approach is able to establish a balance between the prediction and causation and produce clinically more plausible results.



\subsection{Evaluating the multivariate causal hypotheses}
We evaluate the performance of the proposed causal hypothesis generation against the case when we do not use any causal regularization. We generate two lists of top 30 hypotheses using two algorithms and ask our medical expert to label each hypothesis as causal, non-causal or possibly causal with corresponding scores of $1$, $0$, and $0.5$. The results in Figure \ref{fig:hypo_gen} shows that the causal regularizer can increase the causality score of the hypotheses by up to $20\%$.
We also provide a qualitative analysis of the causal hypotheses generated by our algorithm by picking several hypotheses and showing that  they are clinically meaningful. Three examples of multivariate causal hypotheses generated via causal regularizer and the description of their clinical meaning are shown in Table \ref{tab:hypo}.

\section{Conclusion and Discussion}
We addressed the problem of exploring the high-dimensional causal hypothesis space in applications such as healthcare. We designed a causal regularizer that maximally steers predictive models towards causally explainable models. The proposed causal regularizer, based on our causality detector, does not increase the computational complexity of the $L_1$ regularizer and can be seamlessly integrated with a neural network to perform non-linear causality analysis. We also demonstrated the application of the proposed causal regularizer in generating multivariate causal hypotheses. Finally, we demonstrated the usefulness of the causal regularizer in detecting the risk factors of heart failure using an electronic health records dataset.

\section*{Acknowledgment}
The authors would like to thank Frederick Eberhardt for helpful discussions. Mohammad Taha Bahadori acknowledges the previous discussions with David C. Kale and Micheal E. Hankin on the concept of causal regularizer. This work was supported by the National Science Foundation, award IIS-\#1418511 and CCF-\#1533768, research partnership between Children's Healthcare of Atlanta and the Georgia Institute of Technology, CDC I-SMILE project, Google Faculty Award, Sutter health, UCB and Samsung Scholarship. Krzysztof Chalupka's work was supported by the NSF grant \#1564330".

\bibliography{references}
\bibliographystyle{chicago}

\newpage
\clearpage

\appendix
\appendix
\section{Proof of the theorem}
\label{sec:proof}
The proof is established based on two results: First, establishing the connection between the two estimates and the ordinary least squares estimate and then using the known asymptotic normality results for maximum likelihood estimation under misspecification. We need the latter, because the noise distribution is not necessarily Gaussian in the theorem to allow causality detection.

We can write both of the $L_2$ based causal regularization and Ridge regression in the following unified format:
\begin{equation*}
\widehat{\bm{\beta}} = \argmin_{\bm{\beta}} \left\{\frac{1}{n} \left\|\mathbf{y} - \mathbf{X}\bm{\beta}\right\|_2^2 + \lambda\|D^{1/2}\bm{\beta}\|_2^2 \right\},
\end{equation*}
where $D$ is a diagonal matrix $\mathrm{diag}(d_1, d_2)$. It is equal to identity matrix $(d_1, d_2) = (1, 1)$ for the ridge regression and $(d_1, d_2) = (\epsilon, 1-\epsilon)$ for the causal regularization. Given this unified formulation, we can represent the estimates in the theorem in terms of the ordinary least squares estimate as follows:
\begin{align}
\widehat{\bm{\beta}}(D, \lambda) &= (\Xb^{\top}\Xb + \lambda D)^{-1}\Xb^{\top}\yb, \nonumber\\
& = (I + \lambda D)^{-1}\Xb^{\top}\yb, \nonumber\\
& = (I + \lambda D)^{-1}(\Xb^{\top}\Xb)^{-1}\Xb^{\top}\yb,\nonumber\\
& = (I + \lambda D)^{-1}\widehat{\bm{\beta}}_{\text{OLS}}, \nonumber\\
& = \widetilde{D}\widehat{\bm{\beta}}_{\text{OLS}}. \label{eq:step1}
\end{align}
where $\widetilde{D} = \mathrm{diag}(\widetilde{d}_1, \widetilde{d}_2) = \mathrm{diag}(\frac{1}{1+\lambda d_1}, \frac{1}{1+\lambda d_2})$. According to the asymptotic normality of the quasi-maximum likelihood estimation in \citep{white1982maximum}, as $n\to \infty$, the ordinary least squares estimate $\widehat{\bm{\beta}}_{\text{OLS}}$ is normal with the following distribution :
\begin{equation}
\widehat{\bm{\beta}}_{\text{OLS}} \sim \mathcal{N}\left(\bm{\beta}, \frac{\gamma^2}{n} I\right).\label{eq:step2}
\end{equation}

Given the results in Eqs. (\ref{eq:step1}) and (\ref{eq:step2}), we can find the distributions for the quantities of interest:
\begin{align}
\widehat{\beta}_1 - \widehat{\beta}_2 &= (1, -1)^{\top}\widehat{\bm{\beta}}(D, \lambda),\nonumber\\
& = (1, -1)^{\top}(\widetilde{D}\widehat{\bm{\beta}}_{\text{OLS}}), \nonumber\\
& \sim \mathcal{N}(\widetilde{d}_1\beta_1-\widetilde{d}_2\beta_2, (\widetilde{d}_1^2+ \widetilde{d}_2^2)\frac{\gamma^2}{n}). \label{eq:step3}
\end{align}
where in the last step we have used the results on linear transformation of multivariate normal variables. Now, we can use the result in Eq. (\ref{eq:step3}) to write:
\begin{align}
\mathbb{P}[\widehat{\beta}_1 > \widehat{\beta}_2] &= \Phi\left(\frac{\sqrt{n}}{\gamma}\frac{\widetilde{d}_1\beta_1-\widetilde{d}_2\beta_2}{\sqrt{\widetilde{d}_1^2+ \widetilde{d}_2^2}}\right), \nonumber\\
& = \Phi\left(\frac{\sqrt{n}}{\gamma}\frac{(1+\lambda d_2)\beta_1-(1+\lambda d_1)\beta_2}{\sqrt{(1+\lambda d_1)^2+ (1+\lambda d_2)^2}}\right),
\end{align}
where $\Phi$ is the cdf of the unit Gaussian distribution. Substituting $(d_1, d_2) = (1, 1)$ for the ridge regression and $(d_1, d_2) = (\epsilon, 1-\epsilon)$ for the causal regularization, we obtain the result in the theorem.

\section{Details of causality detector design}
\label{sec:details_detector}

We first describe the sampling process for generating synthetic datasets used for training the causality detector algorithms in Section \ref{sec:sample}. Next, evaluate the impact of the proposed sampling procedure on the quality of causality detection algorithms. Algorithm \ref{blk:training} summarizes the process described in Section \ref{sec:detector}.

\begin{Algorithm}[t]
\fbox{\parbox{\textwidth}{Given the data, perform the following steps:
\begin{enumerate}[leftmargin=5mm]
\item Generate data samples $S_i$ for $i=1, \ldots, n_{\text{train}}$ from $p_{X, Y}$ according to the ten cases in Figure \ref{fig:cases}.
\item  Assign label $y=0$ to the cases in Figures \ref{fig:direct}, \ref{fig:indirect}, \ref{fig:conf_dir} and \ref{fig:conf_indir} and $y=1$ to the rest. 
\item \label{it:train} Train a classifier $f: \mathcal{S} \to [0, 1]$ to classify them as causation (label=1) or not-causation (label=0). Given the fact that this is a synthetic dataset, we know these labels and we can use supervised learning.
\item On the test set, construct the test sample sets and use the classifier in step \ref{it:train} to classify the example.
\end{enumerate}}}
  \vspace{-0.05in}
  \caption{The algorithm for constructing the causality detector. The structure of neural network classifier is given in Appendix \ref{sec:neural_cause}. \label{blk:training}}  
\end{Algorithm}

\subsection{Sampling procedure for count variables}
\label{sec:sample}
\begin{figure}[ht]
\centering
\includegraphics[scale=0.35]{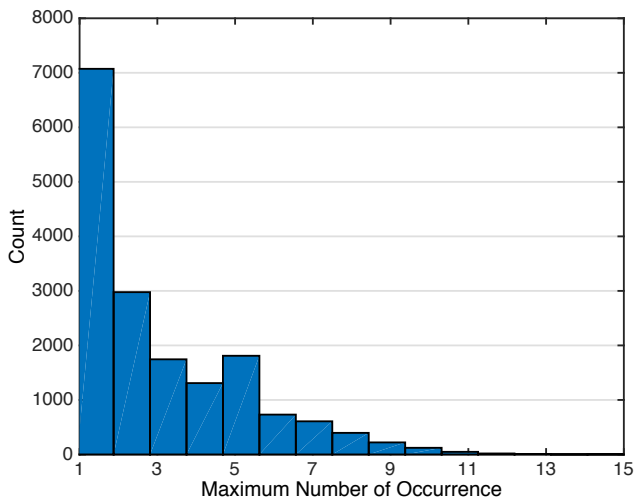}
\caption{Histogram of maximum number of code occurrences.}
\label{fig:maxcount}
\end{figure}

As described in Section \ref{sec:exp}, our independent variables have count data type. Thus, we need to generate data from distributions for count data, such as Poisson or Zipf distributions with fixed support size of 16. Looking at the histogram of maximum number of code occurrences in Figure \ref{fig:maxcount}, we observe that many codes only occur at most once or twice. Thus, we also generate binary and trinary distributions from flat Dirichlet distributions. Finally, to make sure that the space is fully spanned, we also generate samples from Dirichlet distribution with 16 categories. In summary, the $\mathrm{dist}(s, K)$ is the mixture of these five distributions.  The parameters of Poisson and Zipf are sampled from $\chi^2(1)$ distribution.


\begin{Algorithm}[ht]
\fbox{\parbox{\textwidth}{  Let $\mathrm{dist}(s, K)$ denotes a discrete distribution with parameter $s$ and given support size $K$.\\ \vspace{-0.05in}

\textbf{Direct} $X \to Y$:
\begin{enumerate}[leftmargin=7mm]
\item Sample $s\sim \chi^2(2)$. Generate $P_X = \mathrm{dist}(s, K)$.
\item Sample $P_{Y|X} \sim \mathrm{Unif}(0, 1)$ for $K$ times.
\item Compute the $2K$-dimensional vector\\ $\mathbf{P}_{X,Y}(x,y) = [p(1, 0), \ldots, p(K, 0), p(1, 1), \ldots, p(K, 1)]$.
\end{enumerate}}}
  \caption{Another example of generating the synthetic dataset.}  
\end{Algorithm}

Sampling from the other graphical models in Figure \ref{fig:cases} is done writing the factorization and sampling from directed edges and finally marginalization with respect to hidden variables \citep{wainwright2008graphical}. The hidden variables are selected to be categorical variables with cardinality selected uniformly from the integers in the interval $[2, 100]$. The conditional distribution of the hidden variables is selected to be Dirichlet distribution with all-ones parameter vector.

\begin{figure}[ht]
    \centering
    \begin{subfigure}[b]{0.45\textwidth}
        \includegraphics[width=\textwidth]{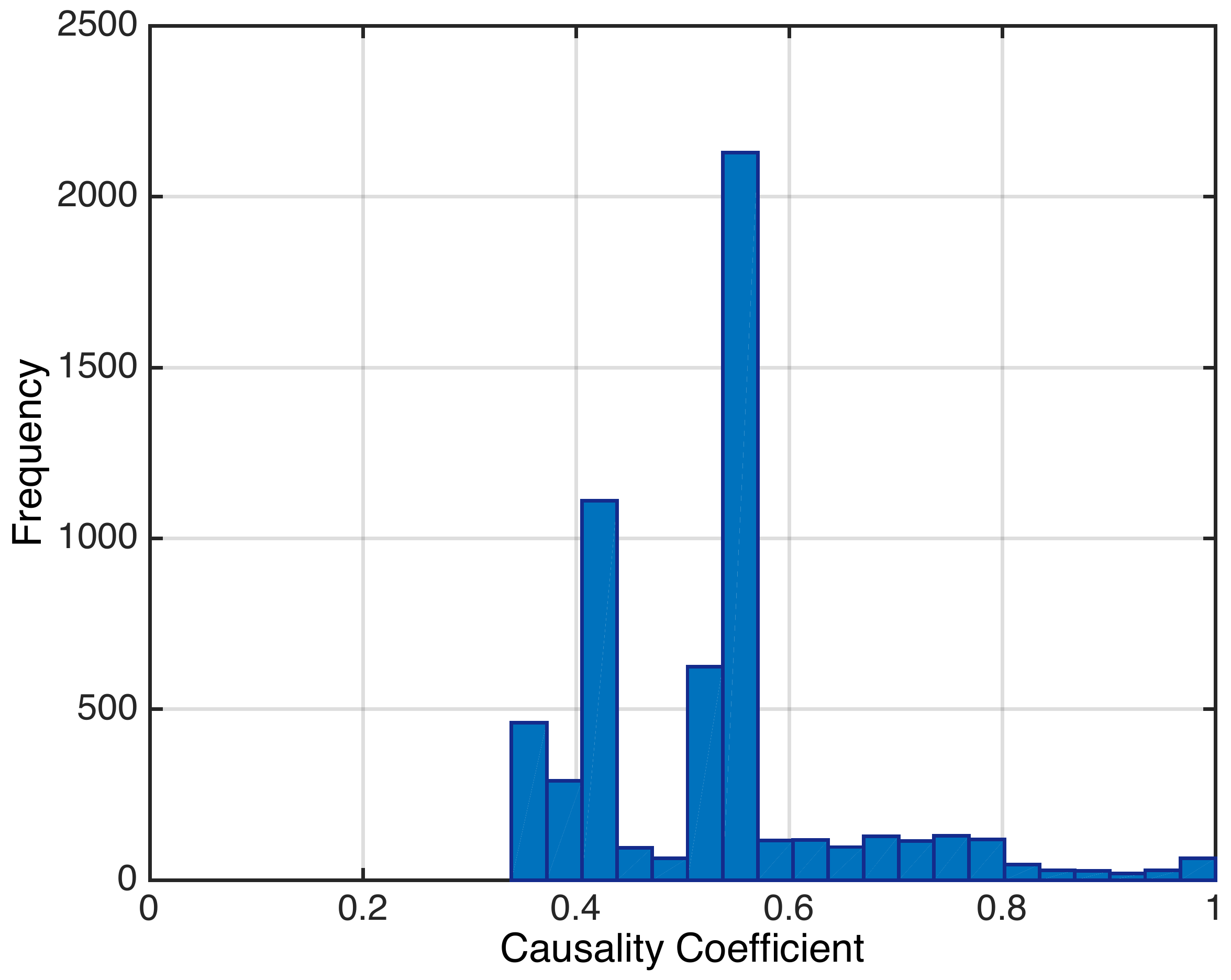}
        \caption{Binary}
        \label{fig:binary}
    \end{subfigure}
    ~ 
    \begin{subfigure}[b]{0.45\textwidth}
        \includegraphics[width=\textwidth]{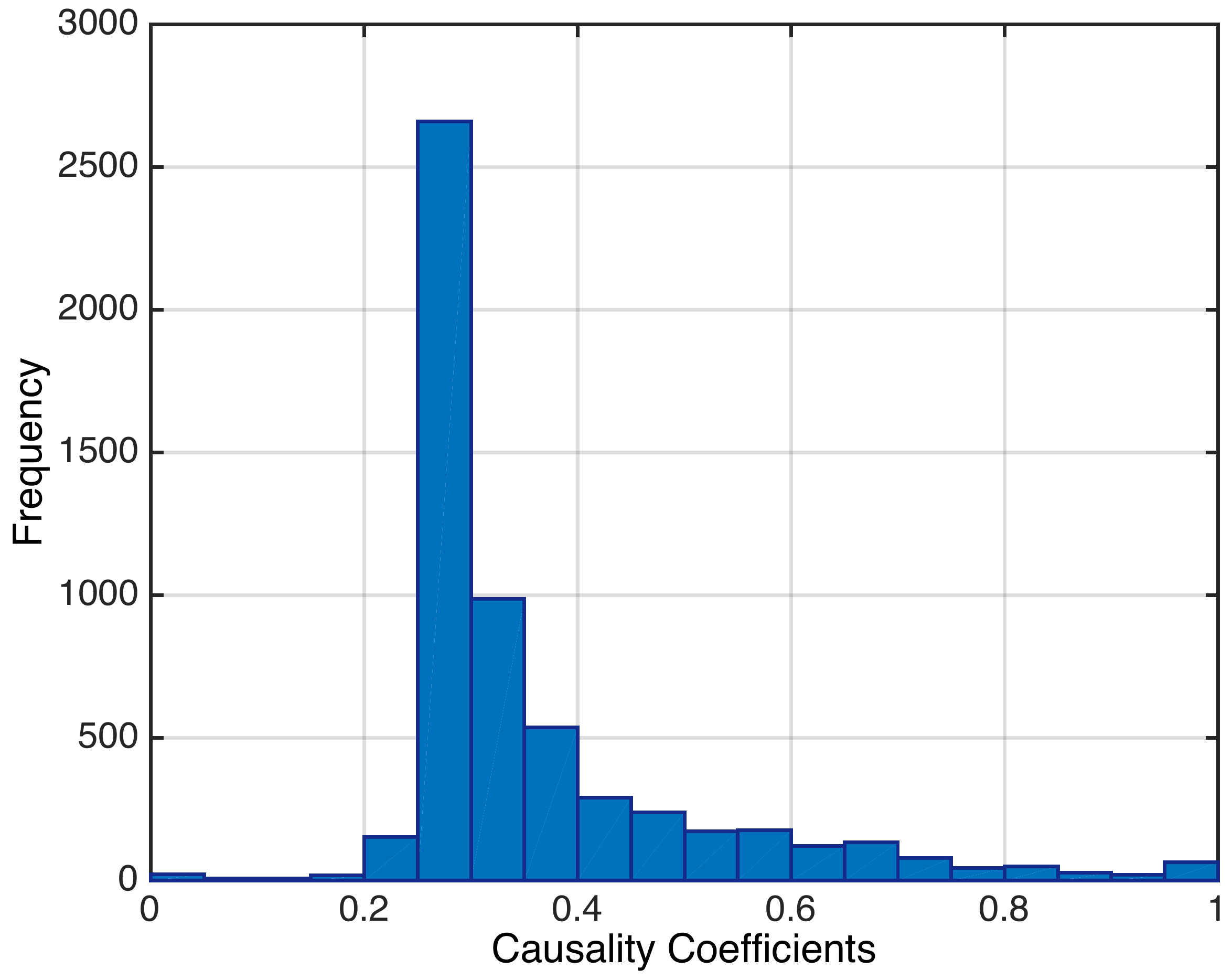}
        \caption{Count}
        \label{fig:count}
    \end{subfigure}
    \caption{Distribution of the coefficients generated by the causality detector.}\label{fig:statistics}
\end{figure}

\subsection{Evaluating the causality detector}
\label{sec:cause_eval}
Table \ref{tab:binary-count} show two advantages of the proposed sampling procedure for count data in comparison to the binary case proposed by \citet{Chalupka2016}. First, in the synthetic dataset, the test error is significantly lower. This is because the size of input to the neural causality detector is $32$ compared to $4$ for the binary case. Applying the causality detectors to our data, we observe that the causality scores generated by our sampling scheme has significantly higher correlation with the mutual information between independent variables and the target label. Figure \ref{fig:statistics}  highlights another advantage of the sampling procedure for count data as it is able to identify a larger portion of the variables as non-causal, which is more in line the expectations. Table \ref{tab:mi_list} shows that the  mutual information identifies V70.0 (Routine general medical examination at a health care facility) as highly correlated, but the causality detector correctly identifies it as non-causal with causality score $0.0000$.

\begin{figure}[ht]
    \centering
    \begin{subfigure}[b]{0.45\textwidth}
        \includegraphics[width=\textwidth]{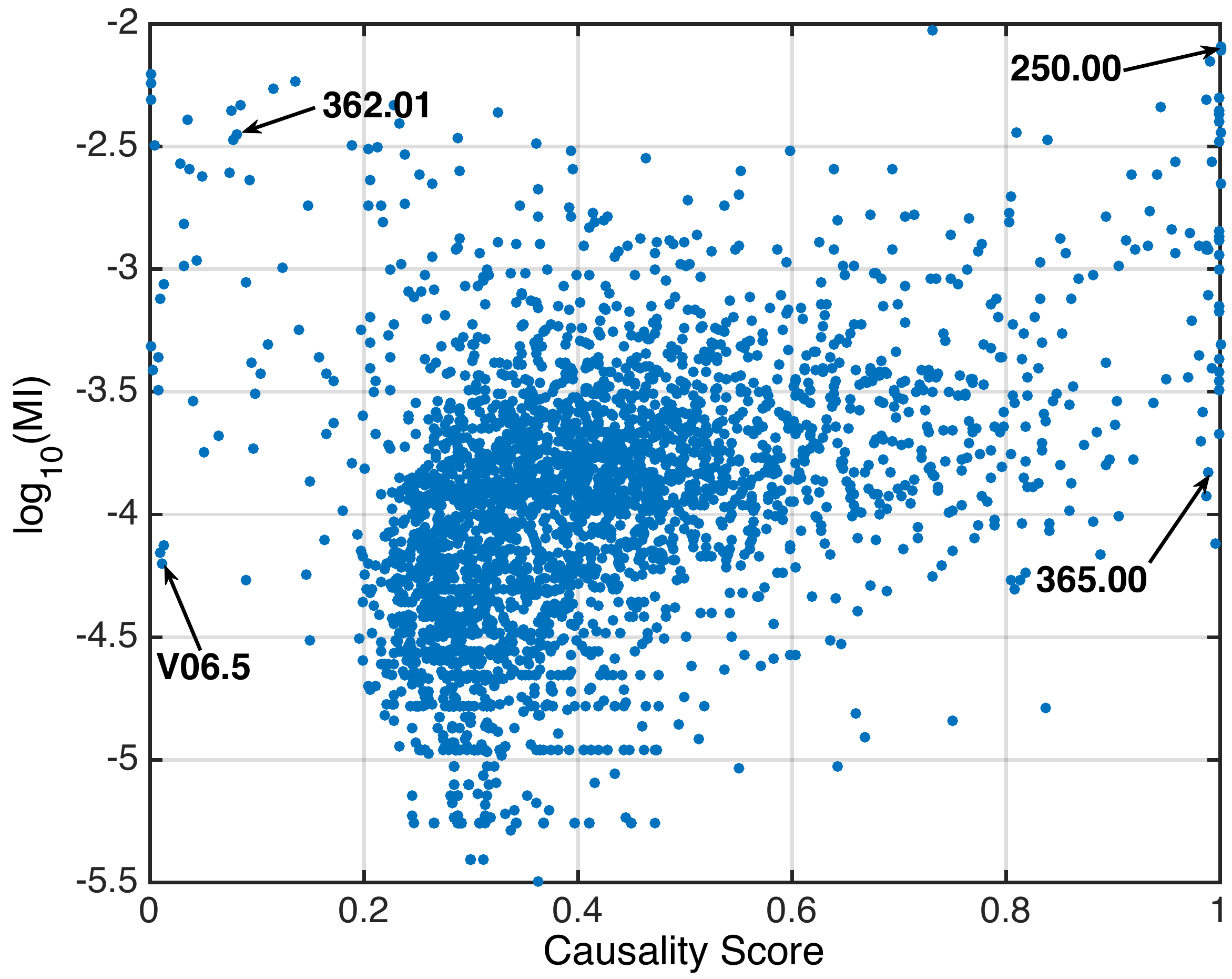}
        \caption{Causation score vs. mutual information}
        \label{fig:mi-cause}
    \end{subfigure}
    \quad 
    \begin{subfigure}[b]{0.45\textwidth}
        \includegraphics[width=\textwidth]{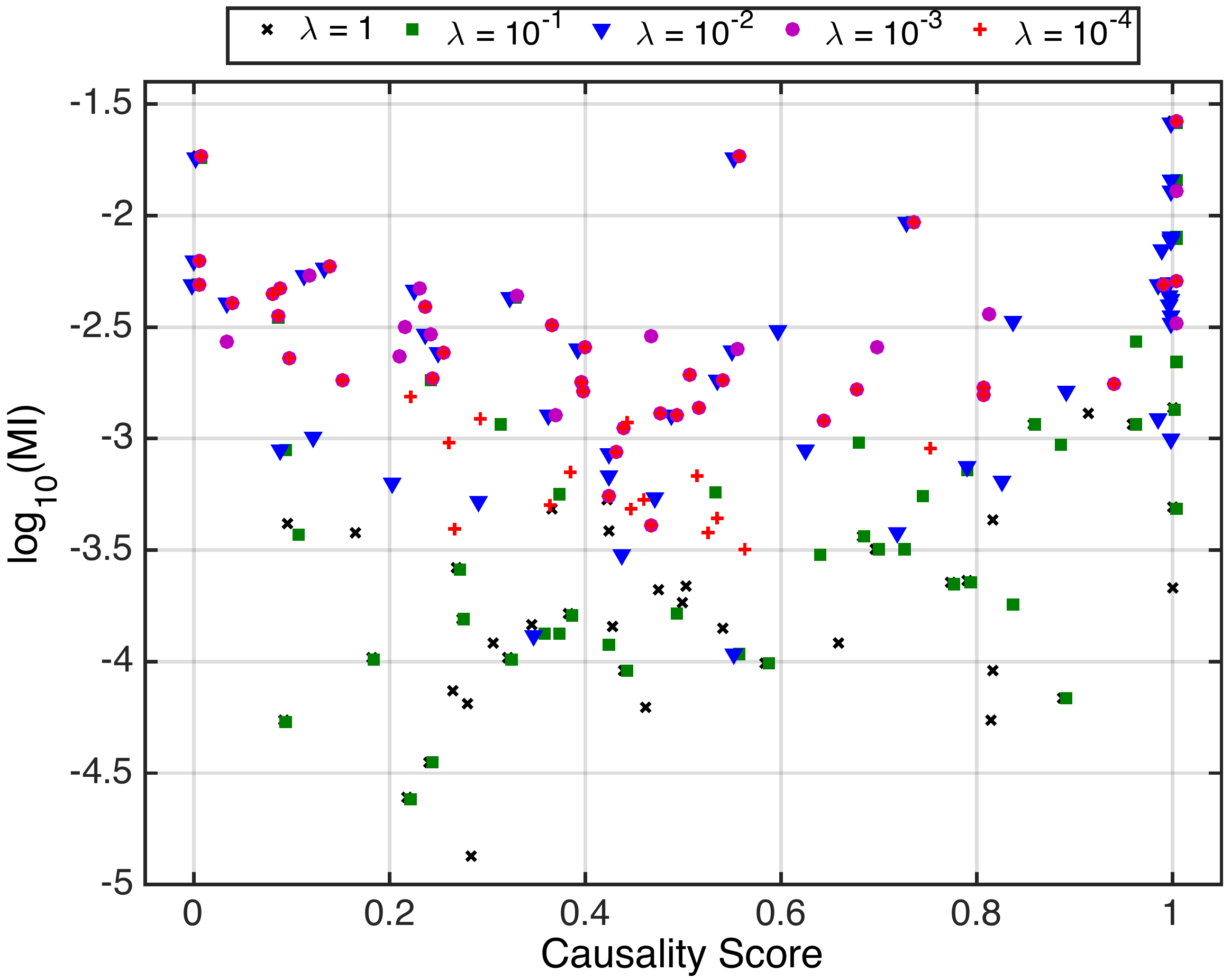}
        \caption{The impact of the causal regularization parameter}
        \label{fig:impact_lamda_vis}
    \end{subfigure}
    \caption{(\subref{fig:mi-cause}) The scatter plot of causation score vs. mutual information. (\subref{fig:impact_lamda_vis}) The impact of $\lambda$ on the top 50 selected variables, marked in the original plot in Figure \ref{fig:mi-cause} As $\lambda$ increases, there is a shift from left-up towards down-right corner and the trade-off shifts towards selecting more causal codes despite possibly lower mutual information. In this figure a small noise has been added to the points to visualize the overlapped points.}
    \label{fig:visualization}
\end{figure}

In particular, in Figure \ref{fig:mi-cause}, the Spearman's rank correlation is $\rho=0.6689$ which indicates a strong correlation. This is intuitive as we expect on average the causal connections to create stronger correlations. Another consequence of the large correlation makes regularization by the non-causality scores safer and guarantees that it will not significantly hurt the predictive performance. In Figure \ref{fig:mi-cause}, we have marked four codes in the four corner of the figure. An example of highly correlated and causal code we can point out 250.00 (Diabetes mellitus without mention of complication) which is a known cause of heart failure. Code 362.01 (Background diabetic retinopathy) is an effect of diabetes ---a common cause of heart failure. Code V06.5 (Need for TD vaccination) is an example of neither causal nor correlated code. Finally, code 365.00 (Preglaucoma) is known for increasing the risk of heart failure, despite the fact that it is not very correlated with heart failure.

\begin{table}[t]
\centering
\caption{Summary of the results \label{tab:binary-count}}
  \begin{tabular}{l|c|c}
\multirow{ 2}{*}{Algorithm} &  Error  & Spearman Correlation \\
 & Rate & w/ Mutual Information \\
\midrule
Binary & 0.2165 & -0.0099 (0.4506) \\
Count & 0.0617 & 0.6689 (0.0000) \\
\end{tabular}
\end{table}

%
%

\section{Details of the neural networks}
\label{sec:neural_cause}
In this section we describe the details of the neural networks used in the paper. Implementation of all methods is done in Theano 0.8 and adamax is used for optimization. We also use early stopping based on the validation accuracy. 

\subsection{Details of CD architecture}
We used a multilayer perceptron with seven layers of size 1024 with rectified linear units as activation functions. We use batch normalization for each layer. 

\subsection{Details of nonlinCause}
\label{sec:nonlineNN}
The $\bm{\alpha}(E\mathbf{x})$ network in the nonlinCause is selected to be a  multilayer perceptron with three layers of size 200 and rectified linear units as activation functions.  Using the described tuning procedure, the embedding dimension $q$ is selected to be $200$ and we used dropout with rate $p=0.8$. The results in Figure \ref{fig:nonlin_gain} is generated using $\lambda=10^{-4}$ for both LogCause and nonlinCause, though we observed similar performance gain for other values of penalization coefficient.

\subsection{Details of CauseHyp}
Implementation of the causality detector in \citep{lopez2016discovering} in our CauseHyp is done via first generating features from the data using a three layered MLP with 200 hidden nodes in each layer. Then, after averaging over the batch, we use a five layered MLP with 200 hidden nodes in each layer. The architecture for the entire network is described in Section \ref{sec:hyp}. 
\balance

\section{Heart failure cohort design}
\label{sec:cohort}
Case patients were 40 to 85 years of age at the time of HF diagnosis. HF diagnosis (HFDx) is defined as: 
\begin{enumerate}
\item Qualifying ICD-9 codes for HF appeared in the encounter records or medication orders. 
\item A minimum of three clinical encounters with qualifying ICD-9 codes had to occur within 12 months of each other, where the date of diagnosis was assigned to the earliest of the three dates.  If the time span between the first and second appearances of the HF diagnostic code was greater than 12 months, the date of the second encounter was used as the first qualifying encounter.  The date at which HF diagnosis was given to the case is denoted as HFDx.
\end{enumerate}
Up to ten eligible controls (in terms of sex, age, location) were selected for each case, yielding an overall ratio of 9 controls per case. Each control was also assigned an index date, which is the HFDx of the matched case. Controls are selected such that they did not meet the operational criteria for HF diagnosis prior to the HFDx plus 182 days of their corresponding case. Control subjects were required to have their first office encounter within one year of the matching HF case patient’s first office visit, and have at least one office encounter 30 days before or any time after the case’s HF diagnosis date to ensure similar duration of observations among cases and controls.

\section{Qualitative results}
\label{sec:tables}
Tables \ref{tab:regu_adv} and \ref{tab:mi_list} are discussed in the experiments section for qualitative evaluation of the results.


\renewcommand{\arraystretch}{0.7}
\begin{table*}[h!]
\centering
\small
\caption{Top 30 codes increasing the heart failure risk identified by the LogCause algorithm. }
\label{tab:regu_adv}
\begin{tabular}{l|p{12.5cm}|c|c}
Code & Description & $\mathbf{w}_{\mathrm{LogCause}}$ & $c_{\text{CD}}$ \\
\midrule
\midrule
794.31& Nonspecific abnormal electrocardiogram [ECG] [EKG]& 0.3422& 0.9351\\
\red{425.8}& \red{Cardiomyopathy in other diseases classified elsewhere}& \red{0.3272}& \red{0.2322}\\
786.05& Shortness of breath& 0.3124& 0.5536\\
\red{424.90}& \red{Endocarditis, valve unspecified, unspecified cause}& \red{0.3086}& \red{0.3908}\\
\red{425.4}& \red{Other primary cardiomyopathies}& \red{0.2880}& \red{0.1351}\\
427.9& Cardiac dysrhythmia, unspecified& 0.2531& 0.9864\\
785.9& Other symptoms involving cardiovascular system& 0.2377& 0.8024\\
585.6& End stage renal disease& 0.2225& 0.3948\\
511.9& Unspecified pleural effusion& 0.2218& 0.0839\\
425.9& Secondary cardiomyopathy, unspecified& 0.2203& 0.8024\\
782.3& Edema& 0.2065& 0.0027\\
\red{278.01}& \red{Morbid obesity}& \red{0.1955}& \red{0.0345}\\
\red{424.0}& \red{Mitral valve disorders}& \red{0.1948}& \red{0.0003}\\
427.31& Atrial fibrillation& 0.1762& 1.0000\\
\red{410.90}& \red{Acute myocardial infarction of unspecified site, episode of care unspecified}& \red{0.1756}& \red{0.2510}\\
426.3& Other left bundle branch block& 0.1690& 0.4890\\
\red{424.1}& \red{Aortic valve disorders}& \red{0.1649}& \red{0.0012}\\
879.8& Open wound(s) (multiple) of unspecified site(s), without mention of complication& 0.1645& 0.6399\\
429.3& Cardiomegaly& 0.1619& 0.5022\\
780.60& Fever, unspecified& 0.1602& 0.7747\\
482.9& Bacterial pneumonia, unspecified& 0.1514& 0.7482\\
786.09& Other respiratory abnormalities& 0.1454& 0.7305\\
496& Chronic airway obstruction, not elsewhere classified& 0.1403& 0.9990\\
V42.0& Kidney replaced by transplant& 0.1398& 0.4351\\
250.03& Diabetes mellitus without mention of complication, type I [juvenile type], uncontrolled& 0.1388& 0.4727\\
276.51& Dehydration& 0.1347& 0.6738\\
403.10& Hypertensive chronic kidney disease, benign, with chronic kidney disease stages I $\sim$  IV& 0.1316& 0.7488\\
\red{250.50}& \red{Diabetes with ophthalmic manifestations, type II, not uncontrolled} & \red{0.1283}& \red{0.2271}\\
427.89& Other specified cardiac dysrhythmias& 0.1282& 0.9416\\
250.51& Diabetes with ophthalmic manifestations, type I [juvenile type], not stated as uncontrolled& 0.1234& 0.5473\\
\end{tabular}
\end{table*}

\begin{table*}[t]
\small
\centering
\caption{Top 20 codes with highest mutual information with the heart failure outcome. $MI$ is the mutual information between $X_i$ and $Y$. }
\label{tab:mi_list}
\begin{tabular}{l|p{12.5cm}|c|c}
Code & Description & $\log(MI)$ & $c_{\text{CD}}$ \\
\midrule
\midrule
782.3& Edema& -1.7355& 0.0027\\
424.1& Aortic valve disorders& -2.2021& 0.0012\\
425.4& Other primary cardiomyopathies& -2.2330& 0.1351\\
\red{V70.0}& \red{Routine general medical examination at a health care facility}& \red{-2.2420}& \red{0.0000}\\
443.9& Peripheral vascular disease, unspecified& -2.2668& 0.1145\\
424.0& Mitral valve disorders& -2.3088& 0.0003\\
250.50& Diabetes with ophthalmic manifestations, type II, not stated as uncontrolled& -2.3288& 0.2271\\
511.9& Unspecified pleural effusion& -2.3320& 0.0839\\
427.32& Atrial flutter& -2.3508& 0.0767\\
278.01& Morbid obesity& -2.3924& 0.0345\\
425.8& Cardiomyopathy in other diseases classified elsewhere& -2.4090& 0.2322\\
362.01& Background diabetic retinopathy& -2.4536& 0.0815\\
584.9& Acute kidney failure, unspecified& -2.4661& 0.2882\\
412& Old myocardial infarction& -2.4750& 0.0780\\
428.0& Congestive heart failure, unspecified& -2.4946& 0.0039\\
791.0& Proteinuria& -2.4984& 0.1883\\
357.2& Polyneuropathy in diabetes& -2.5040& 0.2120\\
402.90& Unspecified hypertensive heart disease without heart failure& -2.5103& 0.2034\\
250.42& Diabetes with renal manifestations, type II or unspecified type, uncontrolled& -2.5310& 0.2378\\
280.9& Iron deficiency anemia, unspecified& -2.5661& 0.0283\\
427.1& Paroxysmal ventricular tachycardia& -2.5884& 0.0367\\
\red{V53.31}& \red{Fitting and adjustment of cardiac pacemaker}& \red{-2.6014}& \red{0.2894}\\
459.81& Venous (peripheral) insufficiency, unspecified& -2.6093& 0.0748\\
410.90& Acute myocardial infarction of unspecified site, episode of care unspecified& -2.6154& 0.2510\\
588.81& Secondary hyperparathyroidism (of renal origin)& -2.6199& 0.0478\\
250.62& Diabetes with neurological manifestations, type II or unspecified type, uncontrolled& -2.6367& 0.2051\\
414.8& Other specified forms of chronic ischemic heart disease& -2.6403& 0.0923\\
362.02& Proliferative diabetic retinopathy& -2.6490& 0.2629\\
586& Renal failure, unspecified& -2.7324& 0.2388\\
250.52& Diabetes with ophthalmic manifestations, type II or unspecified type, uncontrolled& -2.7375& 0.2154\\
\end{tabular}
\end{table*}

\end{document}